\documentclass[10pt,journal,compsoc]{IEEEtran}
\usepackage{amsmath}
\usepackage{graphicx}
\usepackage{subfigure}
\usepackage{color}

\ifCLASSOPTIONcompsoc
  \usepackage[nocompress]{cite}
\else
  \usepackage{cite}
\fi
\ifCLASSINFOpdf
\else
\fi
\begin{document}
%
\title{MulayCap: Multi-layer Human Performance Capture Using A Monocular Video Camera}
%
%
%
%

\author{Zhaoqi~Su, 
        Weilin~Wan, 
        Tao~Yu, 
        Lingjie~Liu, 
        Lu~Fang, 
        Wenping~Wang 
        and~Yebin~Liu
\IEEEcompsocitemizethanks{\IEEEcompsocthanksitem Z. Su, T. Yu, L. Fang and Y. Liu are with Tsinghua University.\protect\\
\IEEEcompsocthanksitem W. Wan, L. Liu and W. Wang are with The University of Hong Kong.\protect\\
\IEEEcompsocthanksitem Corresponding authors: Yebin Liu}}
\IEEEtitleabstractindextext{%
\begin{abstract}
We introduce MulayCap, a novel human performance capture method using a monocular video camera without the need for pre-scanning. The method uses ``multi-layer'' representations for geometry reconstruction and texture rendering, respectively. For geometry reconstruction, we decompose the clothed human into multiple geometry layers, namely a body mesh layer and a garment piece layer. The key technique behind is a Garment-from-Video (GfV) method for optimizing the garment shape and reconstructing the dynamic cloth  to fit the input video sequence, based on a cloth simulation model which is effectively solved with gradient descent. For texture rendering, we decompose each input image frame into a shading layer and an albedo layer, and propose a method for fusing a fixed albedo map and solving for detailed garment geometry using the shading layer. Compared with existing single view human performance capture systems, our ``multi-layer'' approach bypasses the tedious and time consuming scanning step for obtaining a human specific mesh template. Experimental results demonstrate that MulayCap produces realistic rendering of dynamically changing details that has not been achieved in any previous monocular video camera systems. Benefiting from its fully semantic modeling, MulayCap can be applied to various important editing applications, such as cloth editing, re-targeting, relighting, and AR applications.
\end{abstract}

\begin{IEEEkeywords}
Human Performance Capture, 3D Pose Estimation, Cloth Animation, Non-rigid Deformation, Intrinsic Decomposition.
\end{IEEEkeywords}}

\maketitle

\IEEEdisplaynontitleabstractindextext

%
\IEEEpeerreviewmaketitle

\begin{figure*}[ht]
    \includegraphics[width=\linewidth]{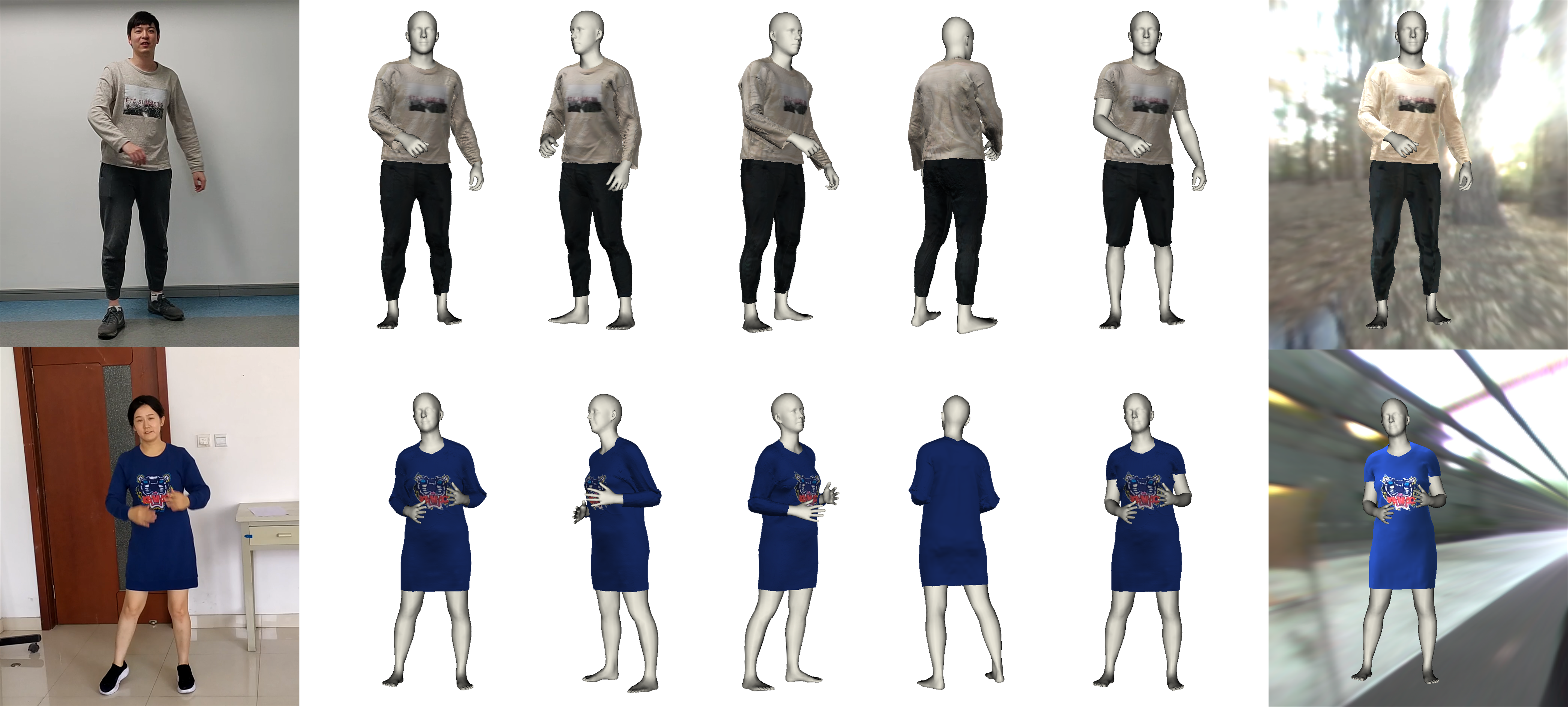}
    \caption{Results generated by our MulayCap system from a monocular RGB video. From left to right: one of input images, four  generated results (one in the reference view and three in different viewing directions), a cloth editing result, and a relighting result rendered under a novel lighting condition.}
    \label{fig:teaser}
\end{figure*}

\IEEEraisesectionheading{\section{Introduction}\label{sec:introduction}}

\IEEEPARstart{H}{uman} performance capture aims to reconstruct a temporally coherent representation of a person's dynamically deforming surface (i.e., 4D reconstruction). Despite the rapid progress in the study on 4D reconstruction using multiple RGB cameras or single RGB-D camera, using a single monocular video camera for robust and accurate 4D reconstruction remains an ultimate goal because it will provide a practical and convenient way of human performance capturing in general scenarios, thus enabling the adoption of human performance capturing technology in various consumer applications, such as augmented reality, computer animation, holography telepresence, biomechanics, virtual dressing, etc. However, this problem is highly challenging and ill-posed, due to the fast motion, complex cloth appearance, non-rigid deformations,  occlusions and the lack of depth information. 

Due to these difficulties, there have been few attempts using a single monocular RGB camera for human performance capture. The most recent works in ~\cite{MonoPerfCap} and \cite{ReTiCaM} approach the problem by using a pre-scanned actor-specific template mesh, which requires extra labor and time to scan, making these methods hard to use for consumer applications or for human performance reconstruction using Internet videos. Moreover, these methods suffer from the limitation of using a single mesh surface to represent a human character, that is, the visible part of human skin and dressed cloth are not separated. As a consequence, common cloth-body interaction, such as layering and sliding, is poorly tracked and represented. Furthermore, once obtained from the pre-scanned template mesh, the reconstructed texture is fixed to the mesh over all the frames, resulting in unrealistic artifacts. 

Without a pre-scanned model, human performance capture is a very difficult problem indeed, due to the need for resolving motion, geometry and appearance from the video frames simultaneously, without any prior knowledge about geometry and appearance. Regarding geometry, reconstruction of a free-form deformable surface from a single video is subject to ambiguity \cite{Dense-non-rigid}. As for texture, it is hard to acquire a dynamic texture free of artifact. Specifically, complex non-rigid motions introduce spatially and temporally varying shading on the surface texture. Directly updating the observed texture on the garment template to represent the motion may introduce serious stitching artifacts, even with ideal and precise geometry models. While artifact-free texture mapping can be obtained by scanning a static key model followed by deforming it in a non-rigid manner for temporal reconstruction, the resultant appearance tends to be static and unnatural.

In this paper, we propose \textit{MulayCap}, a multi-layer human performance capture approach using a monocular RGB video camera that achieves dynamic geometry and texture rendering without the need of an actor-specific pre-scanned template mesh. Here the 'Mulay' notation means that ``multi-layer'' representations are proposed for reconstructing geometry and texture, respectively. We use multi-layer representation in geometry reconstruction, which decomposes the clothed human into multiple geometric layers, namely a naked body mesh layer and a garment piece layer. In fact, two garment layers are used, one for the upper body clothing, such as a T-shirt, and the other for pants or trousers. The upper body clothing can also be generalized to include lady's dresses, as shown in Fig. \ref{fig:teaser}, which uses the same 2D garment patterns as T-shirt, shown in Fig.~\ref{fig:cloth2d3d}, only different in parameters. To solve the garment modeling problem, we propose a Garment-from-Video (GfV) algorithm based cloth simulation. Specifically, the garment shape parameters serve as parameters for cloth simulation and optimized by minimizing the difference between simulated cloth model and the dressed garment observed in the input video. During optimization, to avoid the exhaustive and inefficient search for garment parameters, we use gradient descent minimization with a specified number of iterations. To further align the cloth simulation results with the input images, we apply a non-rigid deformation based on the shape and feature cues in each image. We demonstrate that our proposed garment simulation and optimization framework is capable of producing high quality and dynamic geometry details from a single video. 

The multi-layer representation also works for dynamic texture reconstruction, in which the input video images are decomposed into albedo layers and shading layers for generating albedo atlas with geometry details on clothing. Specifically, each input image is first decomposed into an albedo image and a shading image, then the per-frame albedo is fused with the reconstructed garments to create a static and shading-free texture layer. The albedo layers serve to maintain a temporally coherent texture basis. To obtain a realistic dynamic shape of cloth, we use the shading image to solve for the garment geometry detail with a shape-from-shading method. Finally, by compositing the detailed geometry, albedo and lighting information, we produce high quality and dynamic textured human performance rendering, which preserves the spatial and temporal coherence of dynamic textures and the detail of dynamic wrinkles on clothes. 


In a nutshell, we present a novel template-free approach, called {\em MulayCap}, for human performance capture with a single RGB camera. The use of the multi-layer representations enables more semantic modeling of human performance, in which the body, garment pieces, albedo, shading are separately modeled and elaborately integrated to produce high quality realistic results. This approach takes full advantage of high level vision priors from existing computer vision research to yield high quality reconstruction with light-weight input. In contrast with the existing human performance capture systems~\cite{RelightableHuman,AnimatableHuman,Pons-Moll:2017:ClothCap}, our fully-semantic cloth and body reconstruction system facilitates more editing possibilities on the reconstructed human performances, such as relighting, body shape editing, cloth re-targeting, cloth appearance editing, etc., as will be shown later in the paper. 





%
%
%
%


\section{Related Work}

While there are a large number of prominent works in human performance capture, we mainly review the works that are most related to our approach. We also summarize other related techniques including human shape and pose estimation as well as cloth simulation and capture. 

\textit{Human Performance Capture} The research on Human performance capture has been well studied for many decades in computer vision and computer graphics. Most of the existing systems adopt generative optimization approaches, which can be roughly categorized into multiview-RGB-based methods and depth-based methods according to the capture setups. On the other hand, based on the representations of the captured subject, generative human performance capture methods can be classified as free-form methods, template-based deformation methods and parametric-model-based deformation methods.

For multiview-RGB-based human performance capture methods, earlier researches focus on free-form dynamic reconstruction. These methods use multiview video input by leveraging shape-from-silhouette \cite{IBVH,MustafaKGH15}, multiview stereo \cite{StarckCGA07,LiuTVCG2010,WaschbuschWCSG05} or photometric stereo \cite{VlasicPBDPRM09} techniques. \cite{4DVT_2014_EG} performs video-realistic interactive character animation from a 4D database captured in a multiple camera studio. 
Benefiting from the deep learning techniques, recent approaches try to minimize the number of used cameras (around 4) \cite{MinimalCam18,SparseViewHaoLi18}. 
Template-based deformation methods need pre-scanned templates of the subjects before motion tracking. They can generate topology consistent and temporally coherent model sequences. Such methods take advantage of the relatively accurate pre-scanned human geometry prior and use non-rigid surface deformation \cite{Aguiar08,CarranzaTMS03} or skeleton-based skinning techniques \cite{Vlasic08,GallSkeleton09,LiuPAMI13,BroxRGC10,On-set-performance} to match the multi-view silhouettes and the stereo cues. There have been few studies focusing on temporally coherent shape and pose capture using monocular RGB video sequences. Existing works include \cite{MonoPerfCap} and ~\cite{ReTiCaM}, where a pre-scanned textured 3D model is a pre-requisite for both of them. In their methods, 3D joint positions are optimized based on the CNN-based 2D and 3D joint detection results. Moreover, non-rigid surface deformation is incorporated to fit the silhouettes and photometric constraints for more accurate pose and surface deformation. In parametric-model-based deformation methods. 
The character specific models used in the methods above are replaced by parametric body models like \cite{SCAPE,LoperMB14,sigal2008combined,HaslerARTS10,PlankersF01,SongTCYTZ16} to eliminate the pre-scanning efforts. 
However, parametric body models always have limited power to describe the real world detailed surface of the subject. 
Overall, as most of the template-based deformation methods regard the human surface as a single-piece of watertight geometry, various free-form garment motion and garment-body interactions cannot be described by the surface deformation, which also acts as a key preventer for high quality dynamic texture mapping.

Depth-based methods are relatively more efficient as the 3D surface point clouds are provided directly. 
Many of the previous works in this field are free-form approaches, in which an in-completed template is gradually fused given continuous depth observations. 
Such free-from methods start from the fusion of a general dynamic scene \cite{newcombe2015dynamic}, and have been improved by considering texture constraints \cite{Innmann16,guo2017real} and resolving topology changes \cite{KillingFusion,SobolevFusion}. Multiple depth sensor based fusion approaches  \cite{collet2015high,orts2016holoportation,dou2016fusion4d,dou2017motion2fusion} have been developed to improve the robustness and accuracy through registering multi-view depth streams.
Besides free-form fusion based methods, performance capturing using template-based deformation is also a well studied area. \cite{li2009robust,guo2015robust,ye2012performance, zollhofer2014real} leverage pre-scanned models to account for non-rigid surfaces, while in \cite{helten2013real,ye2014real,bogo2015detailed,walsman2017dynamic} the performance capturing problem is decomposed into articulated motion tracking and shape adaption. \cite{Zhang_2017_CVPR} builds BUFF Dataset which contains high quality clothed 3D scan sequences of the human, and estimates the human body shape and pose from these sequences. There are also some fusion-based approaches combining articulated templates or articulated priors for robust motion tracking and surface fusion \cite{Yu2017BodyFusion, ArticulatedFusion, DoubleFusion, XuPAMI19, Zheng_2018_ECCV}. 

Recently, benefiting from the success of deep learning, discriminative approaches for single image human shape and pose detection catch lots of research attention. They have demonstrated that it is possible to estimate the human shape and pose using only a single RGB image by taking advantage of the parametric body models \cite{SCAPE,SMPL}. \cite{Bogo:ECCV:2016} optimizes the body shape and pose parameters by minimizing the distance between the detected 2D joints from a CNN-based pose detector and the projected 3D joints of the SMPL model. Follow-up works extend this approach by predicting the 3D pose and shape parameters directly. \cite{pavlakos2018learning} proposes a two-step deep learning framework, where the first step estimates key joints and silhouettes from input images, and the second step predicts the SMPL parameters. \cite{omran2018NBF} estimates SMPL parameters through body part segmentation. \cite{HMR} uses a 3D regression module to estimate SMPL parameters and weak camera parameters, and it incorporates an adversarial prior to discriminate unusual poses as well. \cite{3DHumanDynamics} uses temporal information to estimate human poses in a video. \cite{SPIN:ICCV:2019} leverages both the idea from \cite{HMR} and \cite{Bogo:ECCV:2016} and combine the structures from both for iteratively optimization of the human model. \cite{SMPL-X:2019} uses a more expressive model SMPL-X for the human face and hands. Besides, There are also some deep learning approaches for estimating the whole human 3D model or frontal depth map from a single image without using parametric models \cite{pifuSHNMKL19,bhatnagar2019mgn,Zheng2019DeepHuman,Zhu_2019_CVPR,Tang_2019_ICCV}.

\begin{figure*}[ht]
    \centering
    \includegraphics[width=\linewidth]{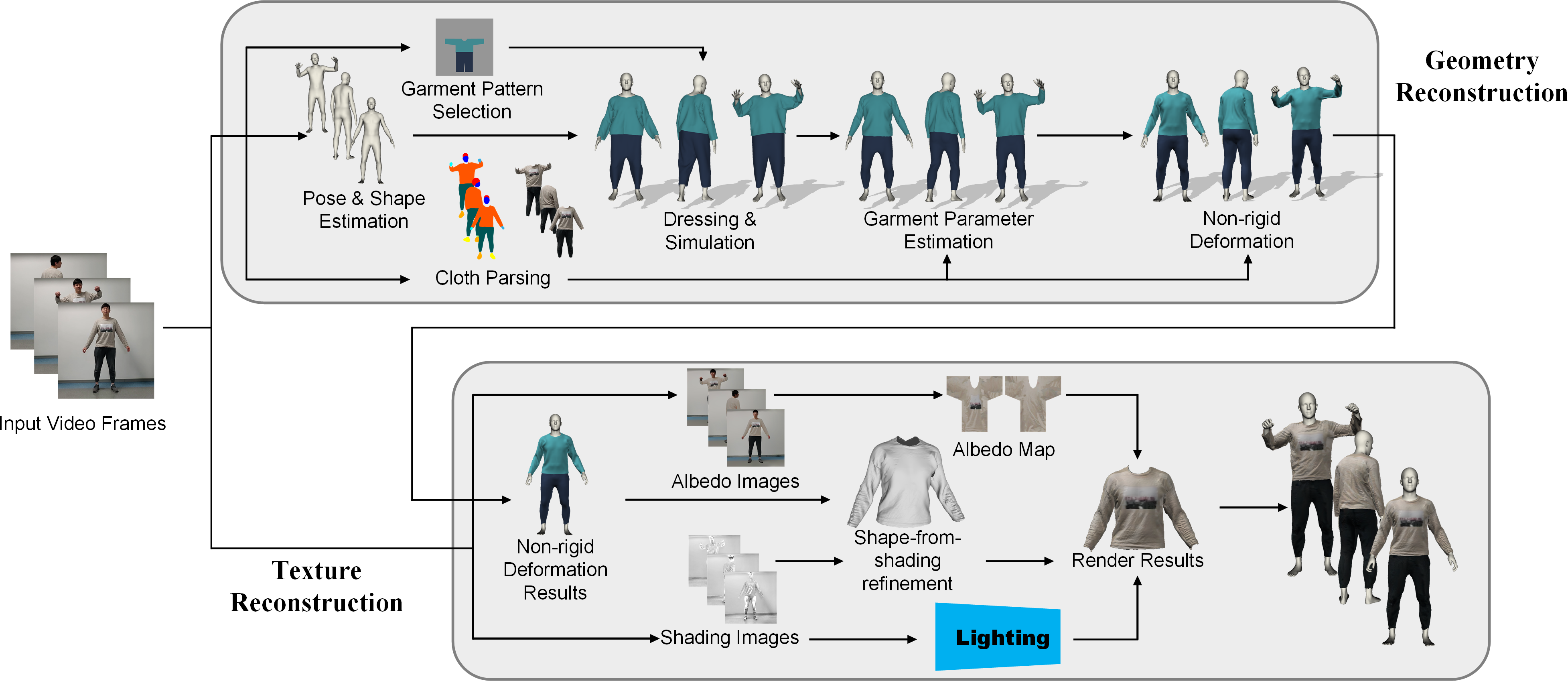}
    \caption{The pipeline of MulayCap. Given the input monocular RGB video, the clothed human reconstruction is achieved by geometry and texture reconstruction. We first estimate the human pose and shape, and reconstruct the garment-based cloth based on the human model, then apply non-rigid cloth deformation based on semantic cloth segmentation result. The second step is to use albedo and shading images decomposed from the input frames to obtain cloth texture, geometry details and lighting, which are then combined for realistic rendering of the dynamic cloth.}
    \label{fig:pipeline}
\end{figure*}

\textit{Cloth Simulation and Capture} \enspace 
The ultimate goal of cloth simulation and cloth capture is to generate realistic 3D cloth with its dynamics. 
Given a 3D cloth model with its physical material parameters, the task of cloth simulation is to simulate realistic cloth dynamics even under different kinds of cloth-object interactions. 
Classical force-based cloth simulation methods are derived from continuum mechanics ~\cite{terzopoulos1987elastically}, it can be a mass-spring system 
~\cite{Provot95deformationconstraints, choi2005stable, liu2013fast} or other more physically consistent models generated by the finite element method 
~\cite{bonet1997nonlinearcontinuummechanics, Jiang:2017:Anisotropic}. These methods need to perform numerical time integration for simulating cloth dynamics, which include the more straightforward explicit Euler method 
~\cite{press2007numerical} and other more stable implicit integration methods like implicit or semi-implicit Euler method 
~\cite{terzopoulos1987elastically, baraff1998large, Bouaziz:2014:Projective}. The force-based cloth simulation methods can generate very realistic cloth dynamics benefiting from the physically consistent models. 
Note that in our MulayCap, cloth simulation is especially useful in dressing the naked body and generating plausible cloth dynamics when only 2D parametric cloth pieces and monocular color video are available. Since the highly accurate material modeling is not a requirement of our system, we use the method in ~\cite{Provot95deformationconstraints} for simplicity and efficiency. 

Different from cloth simulation, the cloth capture methods mainly concentrate on another problem: how to digitize the real world 3D model and even the real world dynamics of the cloth. For active methods, ~\cite{White:2007:CAO:1275808.1276420} custom designed the cloth with specific color patterns and ~\cite{BrostowPAMI11} uses the custom designed active multi-spectral lighting for accurate cloth geometry and even material capture. However, the active methods are much more complex and may not generalize to off-the-shelf clothes. The passive methods are much more popular and have been developed using different kinds of information as input: multi-view rgb ~\cite{Bradley:MarkerlessGarmentCap, Popa:2009, AnimatableHuman}, 4D sequences ~\cite{Pons-Moll:2017:ClothCap,Lahner_2018_ECCV_deepWrinkles}, RGBD ~\cite{Chen:2015:GMD:2816795.2818059,Yu2019SimulCap} or even single RGB ~\cite{garmentFromSingleImage,deepgarment,Yang:2018:PGR:3278329.3026479,alldieck19cvpr,alldieck2019tex2shape,habermann2019TOG,Natsume_2019_CVPR}. 
Among these passive methods, 
~\cite{Bradley:MarkerlessGarmentCap, Popa:2009} focus on reconstructing real cloth geometries and even cloth wrinkle details using temporally coherent multi-view stereo algorithm and data driven approach. 
~\cite{AnimatableHuman} utilizes the multi-view reconstructed 4D human performances to reconstruct a physically-animatable dressed avatar. 
~\cite{Yu2019SimulCap} use a single RGBD camera to accomplish multi-layer human performance capture, which benefits from a physics-based performance capture procedure. 
Given a high quality 4D sequence, ~\cite{Pons-Moll:2017:ClothCap} semantically digitizes the whole sequence and generates temporally coherent multi-layer meshes of both human body and the cloth, while 
~\cite{vivo19garment} propose a multi-task learning framework for garment fashion landmark extraction and garment segmentation from an input image, and can generate garment shape and texture from a single image. 
~\cite{Lahner_2018_ECCV_deepWrinkles} learns a cloth specific high frequency wrinkle model based on normal mapping and use the model to wrinkle the cloth under non-captured poses. 
~\cite{alldieck19cvpr} learns to reconstruct people in clothing with high accuracy, which uses a monocular video of a moving person as input. 
~\cite{habermann2019TOG} build a real-time human performance capture system, which uses RGB video as input and can reconstruct space-time coherent deforming geometry of an entire human. 
~\cite{alldieck2019tex2shape} learns geometry details from texture map, and can infer a full body shape including cloth wrinkles, face and hair from a single RGB image.
~\cite{Natsume_2019_CVPR} uses silhouette information of a single RGB image to infer a textured 3D human model using deep generative models. 
~\cite{Chen:2015:GMD:2816795.2818059} is a data driven approach, it first reconstructs a static textured model of the subject using RGBD sequence and then performs cloth parsing based on a pre-designed cloth database for static but semantic cloth capture; ~\cite{garmentFromSingleImage} improves ~\cite{Chen:2015:GMD:2816795.2818059} by using only a single image. 
~\cite{deepgarment} learns a specific cloth model from a large amount of cloth simulation results with different bodies and poses, and uses it to infer cloth geometry directly from a single color image. 
Benefiting from the parametric cloth models, ~\cite{Yang:2018:PGR:3278329.3026479} can reconstruct both body shape and physically plausible cloth from a single image. However, such method mainly focuses on static cloth reconstruction, thus the cloth dynamics can only be generated by simulation. While Our method can generate realistic cloth dynamic appearance given a video sequence. 
To conclude, on one hand, the data driven approaches above need either (captured or simulated) high quality 4D sequences or self-designed cloth databases as input, which is hard to obtain. Moreover, the generalization of such approaches remains challenging. On the other hand, current direct cloth capture approaches still need carefully designed setups or the heavy multi-view capture systems for high quality cloth capture. 
In our system, we use a data-driven approach to reconstruct static cloth models and propose a new direct cloth capture approach for capturing realistic cloth dynamic appearance from a monocular video footage. There are also many interesting applications of cloth simulation and capture. For example, ~\cite{garmentdesign_Wang_SA18} proposes a learning method which compiles 2D sketches, parametric cloth model and parametric body model into a shared space for interactive garment design, generation and fitting. 
One interesting work correlates to ours is ~\cite{Rogge:2014:GRM:2702692.2634212}, it mainly focuses on garment replacement (but not capture) given a monocular video footage and relies on manual intervention. However, our method is fully automatic and produces both realistic cloth capture and cloth replacement results.

\textit{Intrinsic Decomposition} \enspace The objective of intrinsic decomposition is to decompose a raw image into the product of its reflectance and shading. Because the decomposition of raw images is insufficiently constrained, optimization based methods often tackle the problem by carefully designed priors \cite{barron2015shape,zhao2012closed,bell2014intrinsic,GarcesMLG12}, while deep-learning based methods incorporate learning from ground truth decomposition of raw images \cite{nestmeyer2017reflectance, janner2017self, li2018learning}. There are also sequences based solutions, using propagation methods \cite{YeIntrinsic,bonneel2014interactive,meka2016live}, or considering reflectance as static while shading changes over time \cite{matsushita2004estimating, laffont2015intrinsic}. Methods proposed in \cite{laffont2013rich,duchene2015multi} leverage multi-view inputs to recover the scene geometry and estimate the environment lighting. 

Intrinsic decomposition has been widely applied for identifying the true colors of objects and analyzing interactions with lights in the scene. Researchers have presented various applications based on the progress in this field, such as material editing by shading modification and recolorization by defining transfer among the origin and the target reflectance \cite{YeIntrinsic}. Since most wrinkles and folds on the cloth mainly contribute to the shading effects of the input frames, intrinsic decomposition can be directly incorporated into our system to recover such details.

\section{Method Overview}

The main pipeline of our MulayCap consists of two modules, i.e., multi-layer based geometry reconstruction (see Sect.~\ref{subsec:geometry}) and multi-layer based texturing rendering (see Sect.~\ref{sec:texturing}), as shown in Fig. \ref{fig:pipeline}. For the geometry module, we reconstruct a clothed human model for each input video frame. Each target clothed model contains separated geometry mesh layers for individual garments and the human body mesh model SMPL~\cite{SMPL}. We first detect and optimize the human shape and pose parameters of SMPL model to get the body layer (see Sect.~\ref{subsec:poseshape}). We select the 2D garment pattern and automatically dress the human temporal body models using available cloth simulation methods~\cite{provot1995deformation} (see Sect.~\ref{subsec:dressing}). After that, since the garments may not fit with the input images, we optimize the 2D garment shape parameters using all the 2D segmented garment pieces obtained by instance human parsing methods like ~\cite{InstanceHumanParsing} from the video frames (see Sect.~\ref{subsec:solveparam}). We name this garment shape optimization method based on cloth simulation as GfV. To further align the boundary in each temporal image, we refine the non-rigid deformation of the garments based on the silhouette information in each input image (see Sect.~\ref{subsec:refinement-non-rigid}). 

For the texture module, to achieve temporally dynamical and artifact-free texture updating, we composite a static albedo layer and a constantly updated geometry detail layer on the 3D garments. The garment albedo layer represents a clean and shadow-free texture while the geometry detail layer describes the dynamically changing wrinkles and shadows. First, based on the obtained clothed model sequence in the geometry module, we leverage the intrinsic decomposition method in \cite{nestmeyer2017reflectance} to decompose the input cloth images into albedo images and shading images. Multiple albedo images are then stitched and optimized on the 3D garment to form a static albedo layer (see Sect.~\ref{subsec:albedo}). For the geometry detail, we further decompose the shading images into environment lighting and surface normal images (see Sect.~\ref{subsec:shading}). The normal images are then used to solve for  surface details on the 3D garments (see Sect.~\ref{subsec:normal}). In this way, by using albedo images to render surface albedo, surface detail and environment lighting, we achieve realistic cloth rendering with temporally varying wrinkle details, without the side effect of stitching texture artifacts. 


\section{Multi Layer Geometry}
\label{subsec:geometry}
To elaborate on the geometry reconstruction in MulayCap, we first describe the reconstruction of body meshes, followed by the dressing-on and optimization of the garment layers. 

\subsection{Body Estimation}
\label{subsec:poseshape} 
We use SMPL \cite{SMPL} to track the human shape and pose in each frame. Specifically, We first use HMMR method~\cite{humanMotionKanazawa19} to estimate initial pose parameters $\mathcal{P}_i$ and shape parameters $\mathcal{S}_i$ for each frame $\mathcal{I}_i$. All the $\mathcal{S}_i$ are averaged to get consistent SMPL shape for the whole sequence. We apply temporal smoothing to the pose parameters of adjacent frames to alleviate errors and jitter effects, and replace those poses with drastic changes by interpolation of their adjacent poses. We also leverage the 2D joints of humans detect by OpenPose~\cite{OpenPoseCao} to further fix the inaccurate pose detected by HMMR~\cite{humanMotionKanazawa19} and estimate the global translation of the human model, by constraining the 2D distance between the projected 3D joint position and the detect one.

\begin{figure}
    \centering
    \subfigure[]{
    \begin{minipage}[t]{0.65\linewidth}
    \centering
    \includegraphics[width=\linewidth]{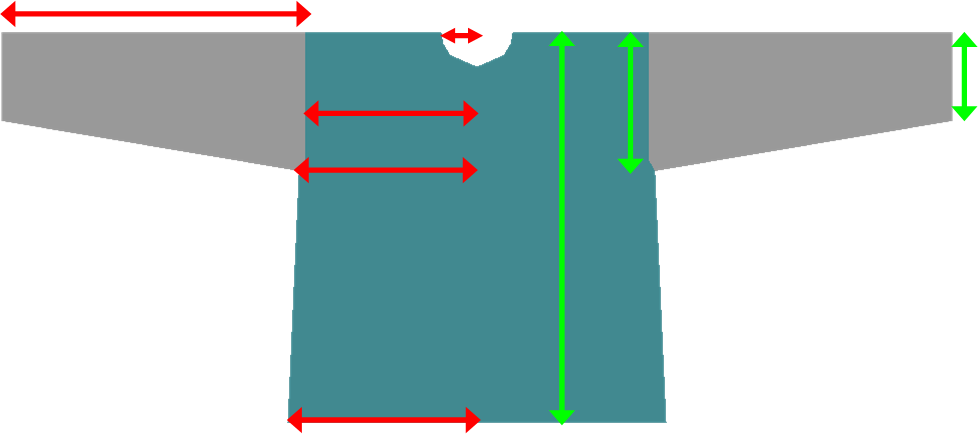}
    \end{minipage}
    }
    \subfigure[]{
    \begin{minipage}[t]{0.30\linewidth}
    \centering
    \includegraphics[width=\linewidth]{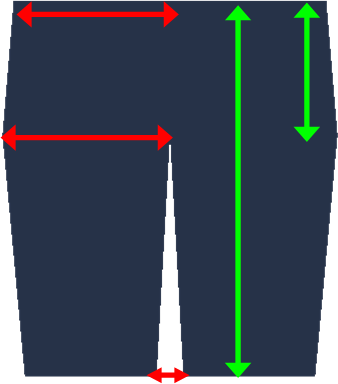}
    \end{minipage}
    }
    \caption{2D garment patterns for generating different clothes. (a) The 2D garment for upper cloth. (b) The 2D garment for pants. The red arrows and green arrows indicate the parameters for controlling the width and height of the clothes, respectively.}
    \label{fig:cloth2d3d}
\end{figure}

\begin{figure*}
    \centering
    \subfigure[]{
    \begin{minipage}[t]{0.17\linewidth}
    \centering
    \includegraphics[width=\linewidth]{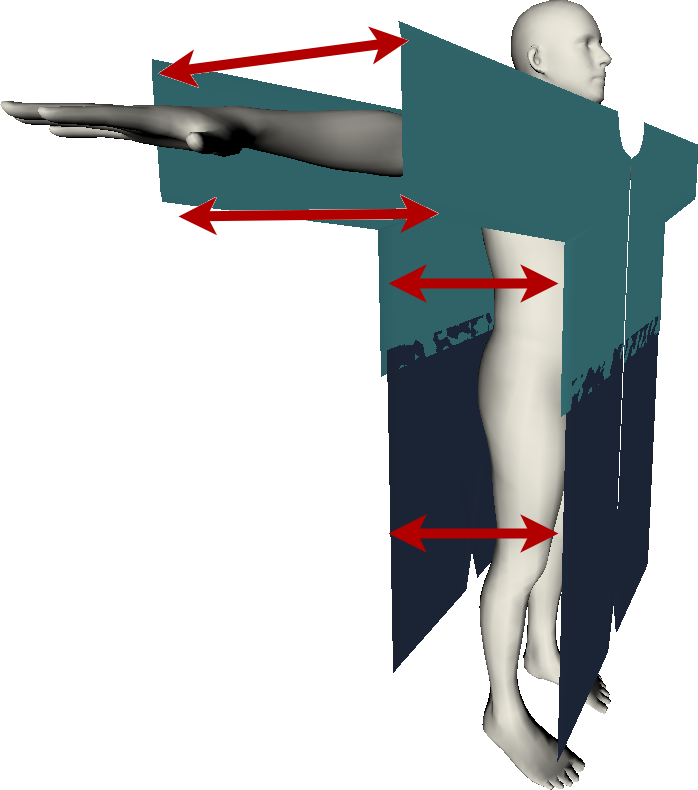}
    \end{minipage}
    }
    \subfigure[]{
    \begin{minipage}[t]{0.17\linewidth}
    \centering
    \includegraphics[width=\linewidth]{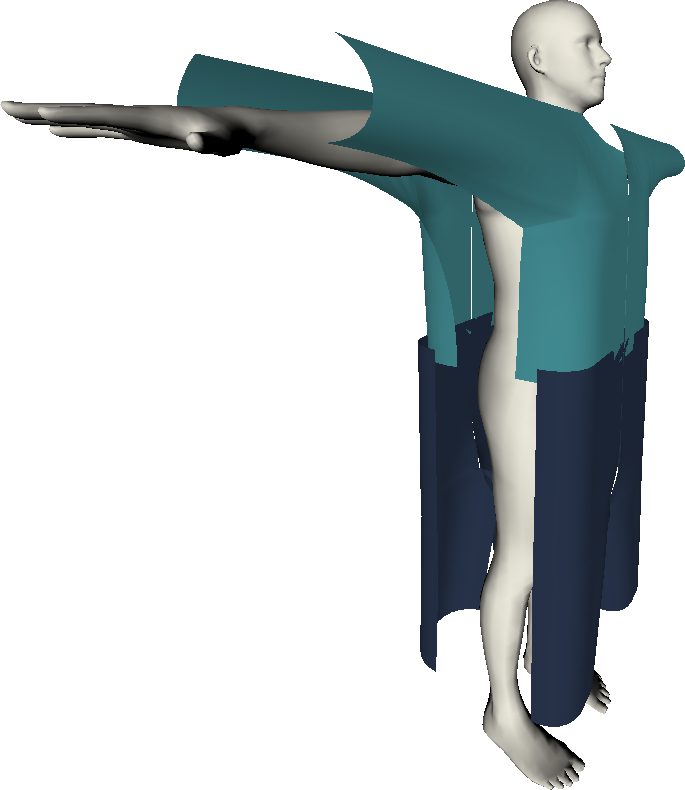}
    \end{minipage}
    }
    \subfigure[]{
    \begin{minipage}[t]{0.16\linewidth}
    \centering
    \includegraphics[width=\linewidth]{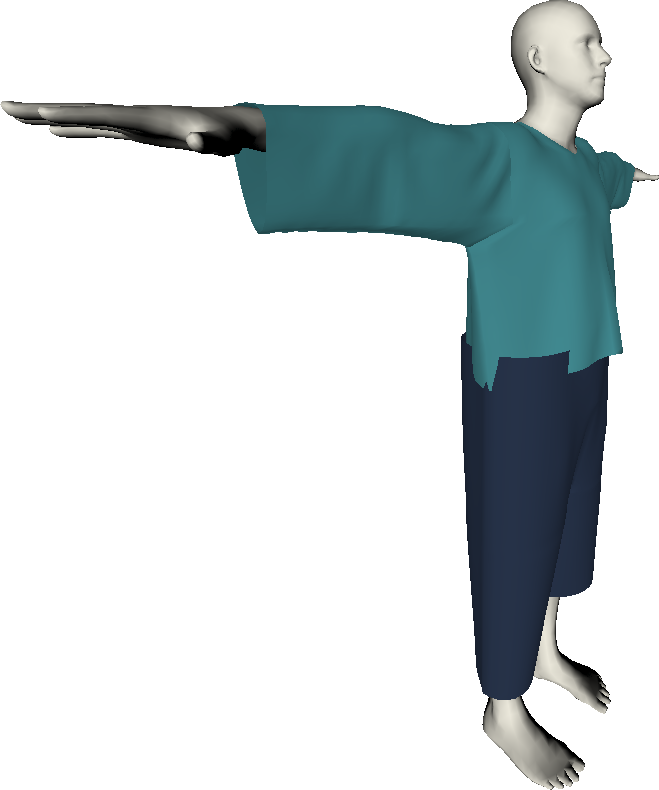}
    \end{minipage}
    }
    \subfigure[]{
    \begin{minipage}[t]{0.16\linewidth}
    \centering
    \includegraphics[width=\linewidth]{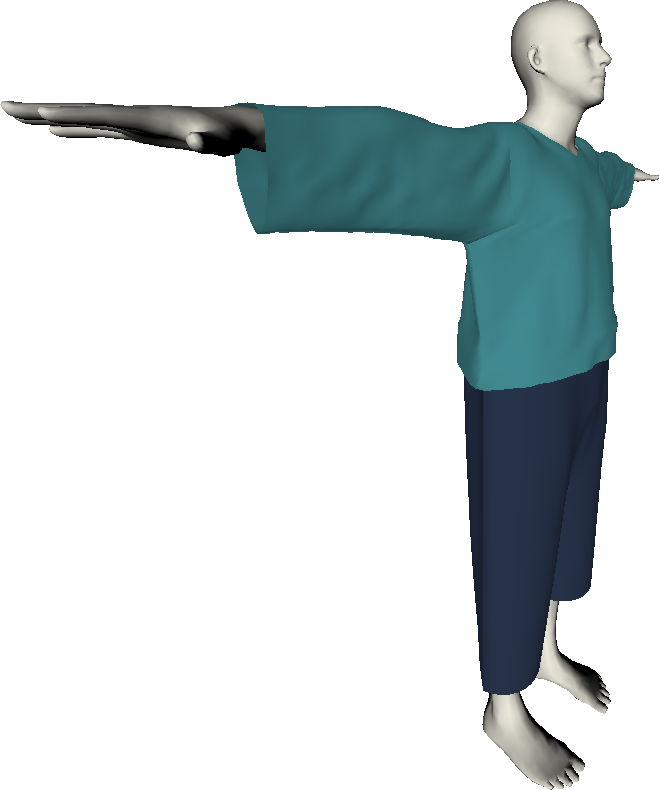}
    \end{minipage}
    }
    \subfigure[]{
    \begin{minipage}[t]{0.12\linewidth}
    \centering
    \includegraphics[width=\linewidth]{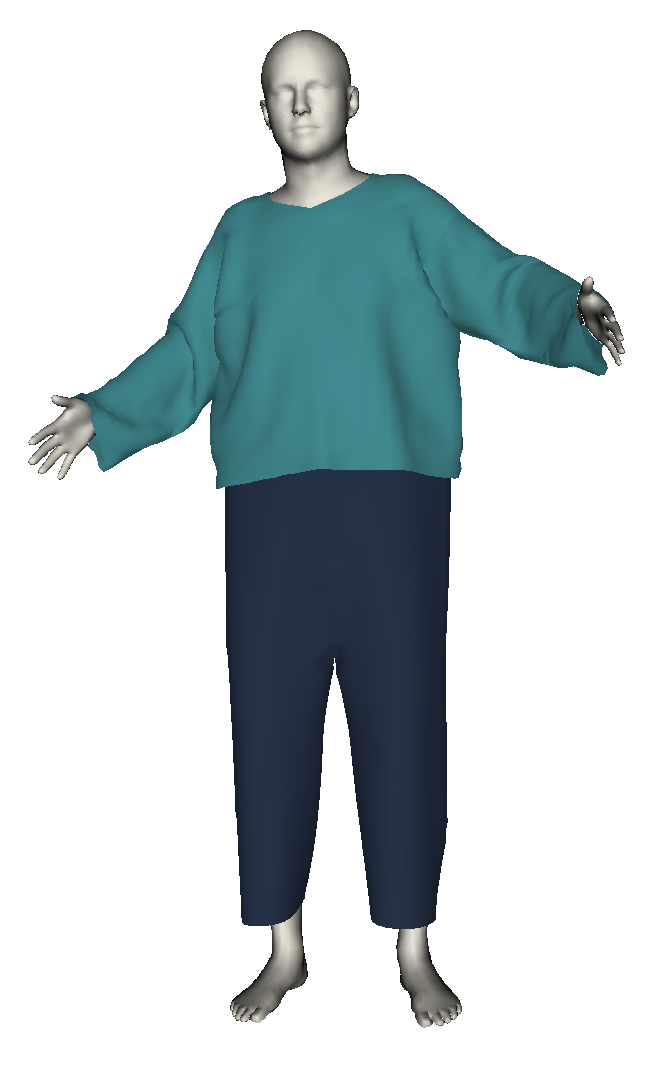}
    \end{minipage}
    }
    \subfigure[]{
    \begin{minipage}[t]{0.12\linewidth}
    \centering
    \includegraphics[width=\linewidth]{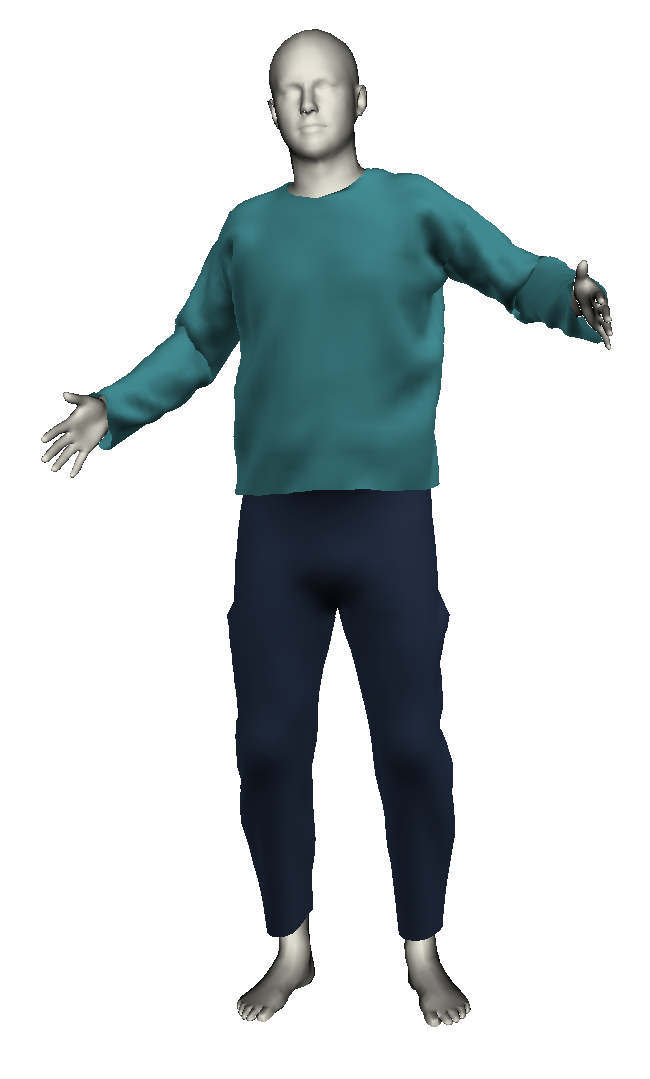}
    \end{minipage}
    }
    \caption{Garment dressing on T-posed body. (a) The initial state of the dressing process, defined by initial garment parameters. The red arrows indicates the attractive force defined on the unseamed garment vertexes. (b) The process of seaming the garment. (c) The cloth is seamed and worn on the body, simulated with gravity and body collision. (d) Resolving garment-garment collision. (e) Simulation result on one of the input frame based on initial garment parameters. (f) Results after the garment parameter optimization under the same pose of (e).}
    \label{fig:seam}
\end{figure*}

\subsection{Garment from Video (GfV)}
\subsubsection{Dressing}
\label{subsec:dressing}
The cloth dressing task consists of two steps: a) garment pattern initialization; b) physical simulation. 
For reconstructing garment layers, multiple 2D template garment patterns are used for different initial 3D garment meshes. Two layers of garments are used for the upper body and the lower body, respectively, as shown in Fig.~\ref{fig:cloth2d3d}. The parameters are defined as the length of the green and red arrows in Fig.~\ref{fig:cloth2d3d}, which leads to 8 parameters for upper cloth and 5 parameters for pants. The parameters are defined in 2D garment patterns, inspired by the industry designing pattern of clothes. Each pattern is composed of a front piece and a back piece. The sizes of garment pieces are specified by the estimated body shape automatically. As shown in Fig.~\ref{fig:seam}(e), the initialization step needs to guarantee the cloth is wide enough to be dressed on. First, we find the average values for all the garment parameters which can fit the average-shape SMPL model. Second, for the given shape, as we want the parameters to roughly fit the body, so that we set the initial width of the cloth according to the bust and waist measurements of the SMPL model with a scale factor of 1.5. Similarly, we utilize the length of the torso and legs of the SMPL model for setting the initial heights for the 2D patterns, for dressing on the body. 
After initialization, the 2D garment parameters can be optimized to fit the real world observations by leveraging both 3D physics-based simulation and numerical differentiation, which is detailed in Sect~\ref{subsec:solveparam}.

To drape the template garments onto a human body mesh in the standard T pose, we introduce external forces to stitch the front and back piece using physics-based simulation, as shown in Fig.~\ref{fig:seam}(a)-(e). The details of cloth dressing and simulation are described below.


For the sake of efficiency, we use the efficient and classical Force-Based Mass-Spring method of  \cite{provot1995deformation} for physics-based simulation, which treats the cloth mesh vertexes as particles connected by springs. For cloth simulation, the external forces applied to each garment vertex include gravity $G$ and the friction between the human body and the cloth. The collision constraints are added between the cloth vertices and the human model, and also the cloth itself (e.g. T-shirt and pants) to avoid inter-penetrations. Specifically, for each cloth vertex $v_i$, we find its nearest SMPL vertex $p_i$ and calculate the point-plane distance $n_i(v_i-p_i)$. If the point-plane distance is below zero, we assume that the collision constraint should be applied to $v_i$. The collision constraints between overlapping cloth working in a similar way. For the whole sequence, the physics-based simulation is conducted sequentially. Specifically, using the draping of the previous frame to initialize the draping of current frame. 

The goal of the dressing step is to seam the two 2D pieces and stitch them as a complete 3D garment. As shown in Fig.~\ref{fig:seam}, to put different garment elements together on the T-posed human body, 
we apply attractive forces on particle pairs at the cutting edges. After about 300 rounds of simulation, a seam detection algorithm is performed to detect whether each unseamed particle-pairs is seamed successfully by detecting whether the distance between the vertex-pairs to be seamed are all smaller than a threshold. 
If not, it means that the initial garment parameters cannot fit the human body, and thus will be automatically re-adjusted. For instance, if the upper cloth cannot be seamed at the wrist part, the algorithm predicts that the cloth is too tight for the human model, and the corresponding garment parameters will be updated progressively until the cloth is successfully seamed, in this case the parameter for sleeve height will be increased gradually until the cloth is seamed well. It should be noticed that, too large garment patterns may deteriorate the dressing process by unstable seaming results. For example, if we set the width for the trouser legs to be too large, during the dressing-on procedure, the two trousers legs will have strong collision effects with themselves, also with the SMPL body, leading to difficulties for generating good seaming results.

\subsubsection{Garment Parameter Estimation}
\label{subsec:solveparam}
After the dressing step, we estimate the garment parameters with several simulation passes. In each pass, the cloth simulation is performed on the whole sequence according to the estimated body shape and poses from Sect.~\ref{subsec:poseshape}. The optimization is initialized with the garment geometry from the initial simulation pass. 

The garment shape is optimized by minimizing the difference between the rendered simulation results and the input image frames. Here we use cloth boundaries in all the image frames as the main constraints for fitting. Given a set of garment shape parameters $\theta_C$, we simulate, render and measure the following energy function $E_{garment}$
\begin{equation}
\label{equ:energy_simulation}
    E_{garment} = E_{bd} + E_{reg},
\end{equation}
where $E_{reg} =||\Delta\theta_C||^2$ is used to regularize the updates between iterations, and $E_{bd}$ is used to maximize the matching between the cloth boundaries. 
For a garment $C$ worn on the body, $E_{bd}$ is defined as
\begin{equation}
\begin{aligned}
    E_{bd} &= \sum_k||\mathcal{F}_k(\theta_{C})||^2 \\
    &\equiv \sum_k||\mathcal{DT}(\mathcal{I}_{img}(C))-\mathcal{DT}(\mathcal{I}_{rn}(\theta_{C}))||^2,
\end{aligned}
\end{equation}
where $k$ is the frame index evenly sampled from the input frames, and $\mathcal{I}_{img}(C)$ is the segmented cloth from the input image obtained with the garment instance parsing method \cite{InstanceHumanParsing},  $\mathcal{I}_{rn}(\theta_{C})$ is the simulated and rendered cloth silhouette using garment parameters $\theta_{C}$, and $\mathcal{DT}(\mathcal{I})$ represents the distance map of the silhouette boundary of image $\mathcal{I}$ and is defined as
\begin{equation}
\label{equ:DT}
    \mathcal{DT}(\mathcal{I}) = max(0, min(\epsilon_{DT}, (\epsilon_{DT}-C(\mathcal{I}))+(\epsilon_{DT}-C(\bar{\mathcal{I}}))),
\end{equation}
where $\epsilon_{DT}$ is a threshold set as 50. Here $C(\mathcal{I})$ and $C(\bar{\mathcal{I}})$ are the distance transform from silhouette of image $\mathcal{I}$ and its inverse image, respectively.


Since the rendering results after cloth simulation are also determined by complex cloth-body interactions, the cloth vertex position cannot be simply formulated as a function of $\theta_{C}$. 
To calculate the gradient of the energy term for Gauss-Newton iteration, we use a numerical differentiation strategy: given the garment parameters $\theta_{C}$, we add a small value $\Delta\theta_{C}^i$ to its $i^{th}$ element, then reset the cloth vertex-pair constraint based on the new garment parameters, and perform the simulation again to calculate the energy for the new parameters as $\mathcal{F}_k(\theta_{C} + \Delta\theta_{C}^i)$. 
The gradient used for Gauss-Newton iteration is then calculated as:
\begin{equation}
    \frac{\partial\mathcal{F}_k(\theta_{C})}{\partial\theta_{C}^i} = \frac{\mathcal{F}_k(\theta_{C} + \Delta\theta_{C}^i) - \mathcal{F}_k(\theta_{C} - \Delta\theta_{C}^i)}{2\Delta\theta_{C}^i}.
\end{equation}

\begin{figure}
    \centering
    \includegraphics[width=0.45\textwidth]{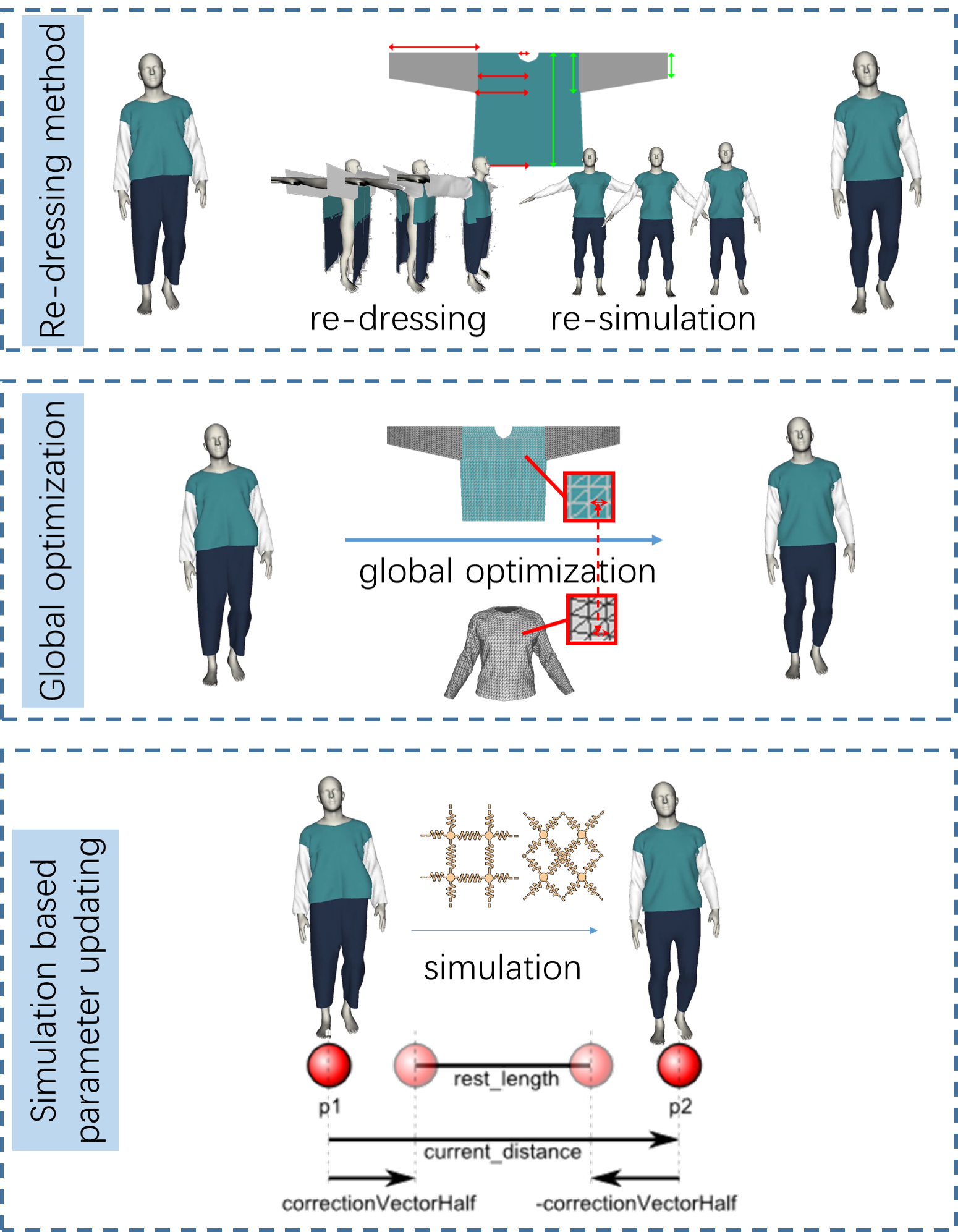}
    \caption{Three ways of performing garment parameter updating: re-dressing method, solver-based global optimization, and simulation-based parameter updating. We perform the third way of parameter updating.}
    \label{fig:3ways}
\end{figure}

To be more specific, when generating 3D garments under new parameters $\theta_{C} + \Delta\theta_{C}^i$ in the $k^{th}$ frame, we first update the corresponding 2D garment patterns, for the garment parameters $\theta_{C}$ are defined on the 2D patterns directly as shown in Fig.~\ref{fig:cloth2d3d}. Generally said, as shown in Fig.~\ref{fig:3ways}, there are three ways of updating the 3D cloth according to the updates of the 2D garment parameters. The most straight-forward way is to re-generate the 2D garment patterns first, and then re-run the cloth dressing step and re-simulate the cloth, which will significantly decrease the efficiency of our method. The second way is to directly update the 3D shape of the cloth according to the changes of the 2D parameters without simulation, which will lead to solving a heavy energy-based global optimization for maintaining the garment structure while constraining the edge length between nearby vertexes to approximate the distance between the corresponding two vertex on the 2D garment pattern. The third way is to leverage the physics-based simulation directly to approximate the parameter changes. In our experiments, we find out that we can directly update the initial simulation status (including force and vertex-pair rest length calculated by 2D patterns) of the garment, and perform the simulation. Specifically, as mentioned before, we use the efficient and classical Force-Based Mass-Spring method for physics-based simulation, which treats the cloth mesh vertexes as particles connected by springs. When the 2D garment parameters are updated, the distance between neighboring vertexes on the 2D garment patterns will be re-calculated, which serves as the new "rest length" between the two vertexes on the cloth mesh. As the "rest length" changes, the vertexes will change its tightness according to the Force-Based Mass-Spring method for physics-based simulation. For example, if the parameter change causes the rest length between vertex $v_1$ and $v_2$ becomes smaller, $v_1$ and $v_2$ will be affected by the attraction force during the simulation, so that the whole garment will change its shape according to the parameter changes during the simulation. In this way, we can generate simulated garment mesh with new parameters without interrupting the simulation or performing the cloth dressing again. 
Finally, after the simulation has reached a stable state under the new parameters, we can use it
to calculate the energy function $\mathcal{F}_k(\theta_{C} + \Delta\theta_{C}^i)$.
The update of $\theta_{C}$ is calculated using Gauss-Newton method. Note that $\Delta\theta_{C}^i$ is only used as a step value for numerical differentiation, which is not the update of $\theta_{C}$ in each iteration. 
The $\Delta\theta_{C}^i$ in our system is set to 0.01.


Our tests show that 25 iterations is generally enough for the above simulation-and-numerical-optimization method to converge and get plausible garment shape parameters.  Fig.~\ref{fig:seam}(f) illustrates the garment shape optimization result over Fig.~\ref{fig:seam}(e).

\subsubsection{Non-rigid Deformation Refinement} 
\label{subsec:refinement-non-rigid}
Note that the garment parameters solved in Sect.~\ref{subsec:solveparam} provides only a rough estimation and the physics-based simulation cannot describe the subtle movement of the cloth, such as wrinkles, which are critical to realistic appearance modeling. We therefore refine the garment geometry using a non-rigid deformation approach to model dynamic cloth details. We up-sample the low resolution cloth mesh used for physics-based simulation to match the pixel resolution for detailed non-rigid alignment and subsequent geometry refinement. Here, we use garment boundary to determine the displacement of each vertex $\Delta{v}_i^h$ in the high resolution garment mesh by minimizing the following energy function
\begin{equation}
\label{equ:energy_highres}
    E_{nonrigid} = E_{bd}^{(h)} + E_{smooth}^{(h)}+E_{reg}^{(h)}.
\end{equation}
Here
\begin{equation}
    E_{bd}^{(h)}=\sum_i ||\mathcal{DT}(\mathcal{I}_{img}({v_i}^h+\Delta{v_i}^h))-\mathcal{DT}(\mathcal{I}_{rn}({v_i}^h))||^2.
\end{equation}
The smoothness term $E_{smooth}^{(h)}$ used to regularize the difference between displacements of the neighboring mesh vertexes:
\begin{equation}
    E_{smooth}^{(h)}=\lambda_{nearby}^{(h)}\sum_i\sum_{j\in\mathcal{N}_i}||\Delta{v_i}^h-\Delta{v_j}^h||^2.
\end{equation}
The regularization term $E_{reg}^{(h)}$ is defined in the same way as $E_{reg}$ to constrain the displacement magnitudes.

\begin{figure}
    \centering
    \subfigure[]{
    \begin{minipage}[t]{0.45\linewidth}
    \centering
    \includegraphics[width=\linewidth]{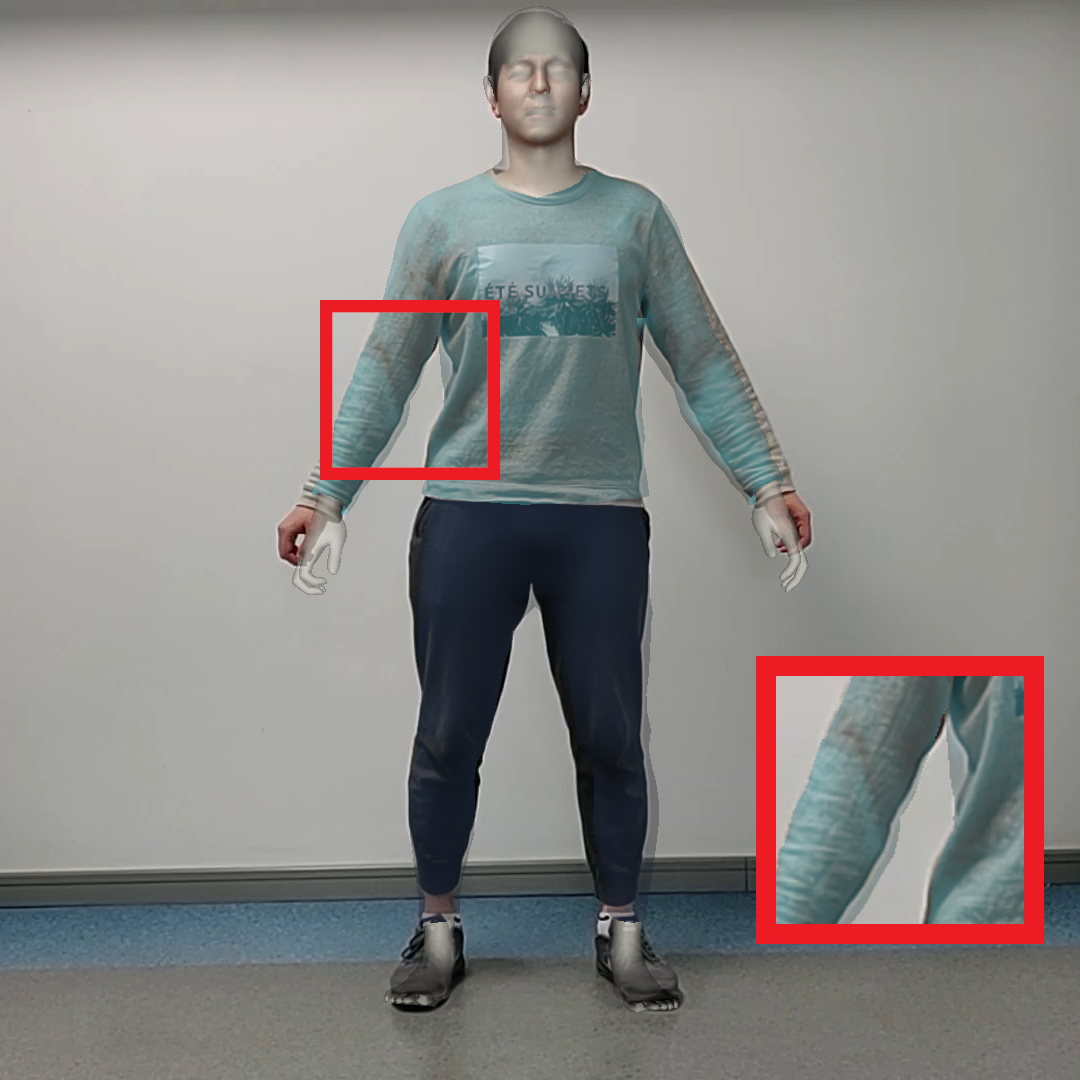}
    \end{minipage}
    }
    \subfigure[]{
    \begin{minipage}[t]{0.45\linewidth}
    \centering
    \includegraphics[width=\linewidth]{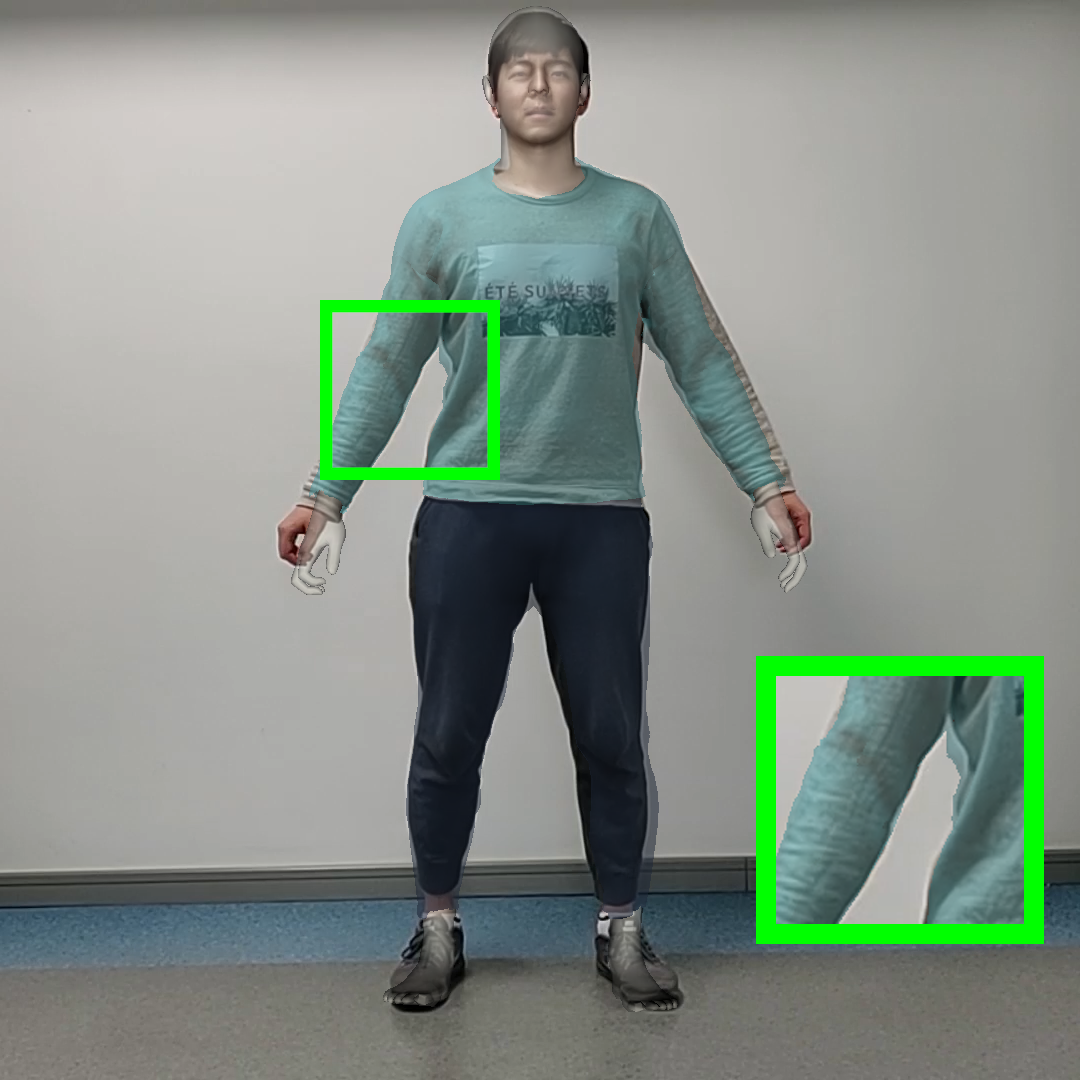}
    \end{minipage}
    }
    \caption{The results of garment deformation refinement. Both (a) and (b) show the reconstructed garment meshes overlay on the input image. Please zoom in to see the overlapping boundaries. (a) The result before deformation refinement. (b) The result after deformation refinement.}
    \label{fig:tracking_solve_deform}
\end{figure}

The energy function in \ref{equ:energy_highres} is minimized using the Gauss-Newton method. 
As the energy term is defined either on each high resolution mesh vertex, or between nearby vertexes, the energy matrix is sparse so that the conjugate gradient algorithm is used in each Gauss-Newton iteration. 
The improvement resulting from refinement by non-rigid deformation is shown in Fig.~\ref{fig:tracking_solve_deform}(a) and Fig.~\ref{fig:tracking_solve_deform}(b). See the zoom-in for detailed boundary overlays. It can be seen that the boundary overlay of rendered mesh with optimization in Fig.~\ref{fig:tracking_solve_deform}(b) is more accurate than that in Fig.~\ref{fig:tracking_solve_deform}(a).  

\section{Texture and geometry detail refinement}
\label{sec:texturing}
In this section we will consider mapping texture to the detailed garment surface reconstructed in the preceding step. Note that directly texturing and updating would introduce serious stitching artifacts due to the spatially and temporally varying shadings and shadows. To obtain artifact-free and dynamically changing surface texture, we decompose the texture into the shading layer and the albedo layer, with the former for geometry detail refinement and global lighting, and the latter for generating a static albedo map for the cloth. 


Specifically, for each input image frame $\mathcal{I}$, we use the CNN-based intrinsic decomposition method proposed in \cite{nestmeyer2017reflectance} to get a reflectance image $\mathcal{I}_{F}$ and a shading image $\mathcal{I}_{S}$. We then use $\mathcal{I}_{F}$ for garment albedo map calculation and $\mathcal{I}_{S}$ for dynamic geometry detail refinement and lighting estimation. All these three components (i.e. detailed garment geometry, albedo map and lighting) are then combined to produce realistic garment rendering.



\subsection{Albedo Atlas Fusion}
\label{subsec:albedo}
To generate a static albedo atlas on each 3D garment, we need to keep a optimized texture base for reducing stitching artifacts and maintaining spatially and temporally consistent texturing. Specifically, for each garment, we build a texture U-V coordinate domain according to the 2D garment designing pattern so that each vertex on a reconstructed garment mesh is assigned to a UV coordinate.

\begin{figure}
    \centering
    \includegraphics[width=\linewidth]{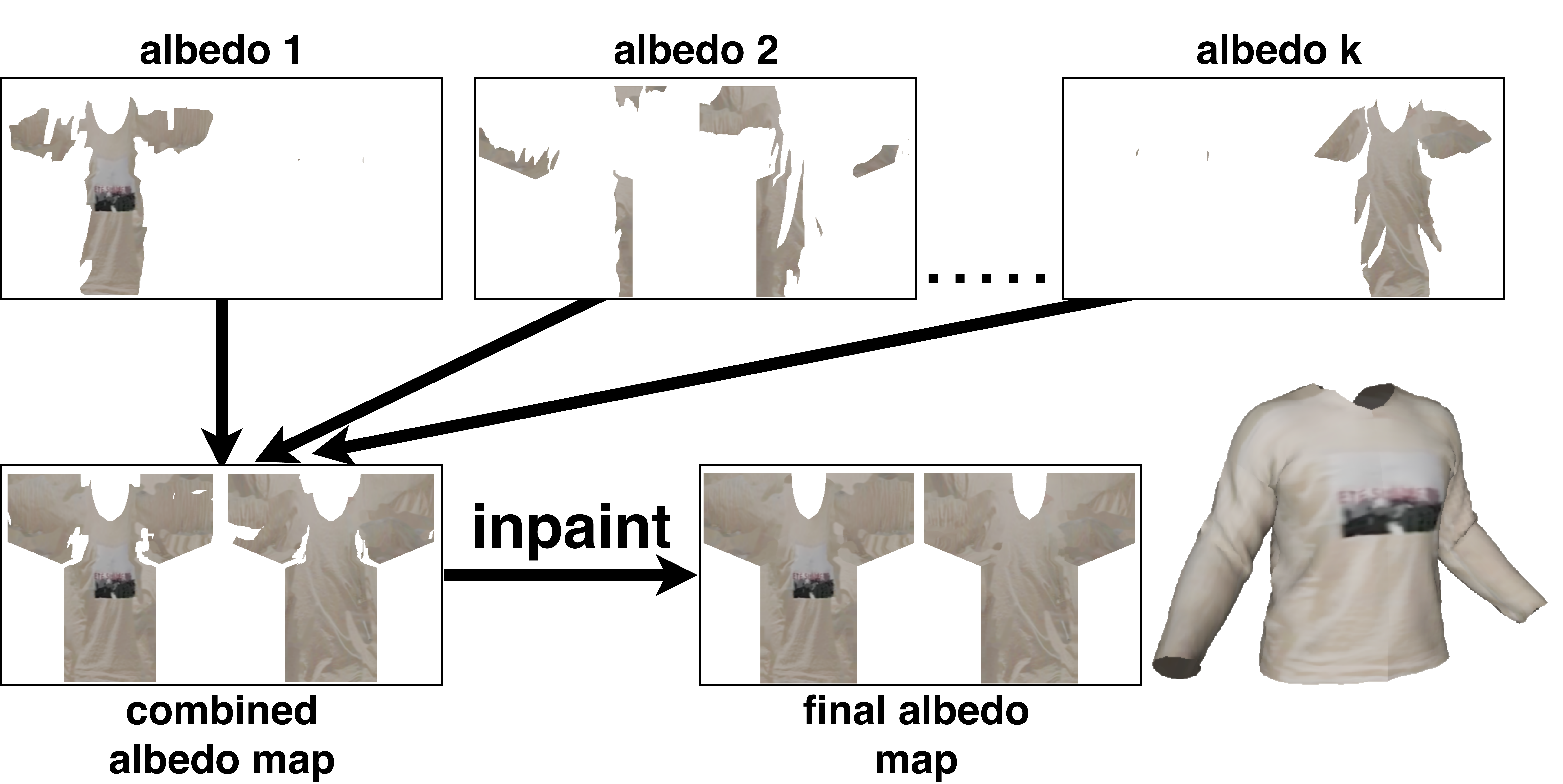}
    \caption{The pipeline of generating the albedo atlas. We iteratively combine albedo maps from evenly sampled key-frames and inpaint the unseen areas.}
    \label{fig:textureCombine}
\end{figure}

Our albedo fusion algorithm creates the albedo atlas based on the evenly sampled albedo images $\mathcal{I}_{F}$. As multiple albedo pixels on different albedo images may project to the same UV coordinate, a multi-image blending algorithm is needed for creating a high quality albedo atlas. 
We resolve this problem by simply using an as-good-as-possible albedo pixel from the multiple images for obtaining the albedo atlas. The selection strategy resembles the multiview texture mapping schemes \cite{eisemann2008floating,buehler2001unstructured}, which try to select the camera that minimizes the angle between the surface normal direction and the vertex-to-camera direction. To mitigate the mosaicing seams, we follow the MRF seam optimization method \cite{lempitsky2007seamless} to remove mosaic seams without affecting the fine details of the albedo. For areas unseen in the sequence, we inpaint \cite{inpaint} those areas to obtain a full albedo atlas. Fig.~\ref{fig:textureCombine} illustrates the albedo atlas fusion pipeline.

\subsection{Shading Decomposition}
\label{subsec:shading}
The goal for shading decomposition is to estimate an incident lighting $\mathcal{L}$ and normal images $\mathcal{I}_{N}$ from the shading image sequence $\mathcal{I}_{S}$. The incident lighting is then used for shape-from-shading based geometry refinement. Following the shading based surface refinement approach in \cite{ChengleiShading}, we use spherical harmonics to optimize the lighting $\mathcal{L}$ and the normal image $\mathcal{I}_{N}$ by minimizing the energy function
\begin{equation}
\label{eqn:shading}
    E_{shading}=\sum_p||\mathcal{H}_M(\mathcal{L},\mathcal{I}_{N}(p))-\mathcal{I}_{S}(p)||^2,
\end{equation} 
which models the different the approximate shading $\mathcal{H}_M$ by spherical harmonics and the observed shading $\mathcal{I}_{S}$ over all the pixels $p$ that belong to a garment in an image frame. 

We get a normal map from the recovered garment surface for lighting optimization, which is a better initialization than uniform normal map initialization. To improve accuracy, we select multiple key frames to enrich the variance of surface normals as done for the albedo atlas generation step in Sect. \ref{subsec:albedo}, and estimate the lighting using the least square method over all the pixels in these frames. We select key frames in an iterative manner. Specifically, if the pose difference between the current frame and all the previously selected key frames is larger than a threshold, we add the current frame as a new key frame; the iteration is stopped when no new key frame can be added. In our experiments, we mainly focus on the torso movements and the threshold is set to be 0.5 for average angle-axis Euler distance among torso joints defined by SMPL model. To constrain the range of normal estimation, we regularize $E_{shading}$ and minimize the energy function $E_{normal}$ based on the estimated lighting $\mathcal{L}$ by minimizing the following energy
\begin{equation}
\label{eqn:energy_pixel_normal}
    E_{normal} = E_{shading} + E_{Lap} + E_{grad} + E_{norm}.
\end{equation}
Here:

regularization term
\begin{equation}
    E_{grad}=\lambda_{grad}\sum_p||\Delta\textbf{n}(p)||^2
\end{equation}
is used to constrain the updating step.

Laplacian term
\begin{equation}
    E_{lap}=\lambda_{lap}\sum_p||Avg_{p'\in N_p}\textbf{n}(p')-\textbf{n}(p)||^2
\end{equation}
is used to constrain the smoothness of the normal image, where $N_p$ is the set of $p$'s neighbor pixels.

normalization term
\begin{equation}
E_{norm}=\lambda_{norm}\sum_p||\textbf{n}(p)^T\textbf{n}(p)-1||^2
\end{equation}
is used to constrain the normal at every cloth pixel $p$ to be normalized.

\subsection{Geometry refinement using shape-from-shading}
\label{subsec:normal}
So far, we have obtain the incident lighting of the scene. Then we use the shape-from-shading approach to refine the geometry detail using both lighting and shading image sequence $\mathcal{I}_{S}$ to represent wrinkles and folds for realistic rendering. Benefiting from the proposed multi-layer representation and physics-based optimization, we can obtain more reasonable garment geometry as the initial status for this geometric detail refinement step. To compute the per-vertex displacement of cloth, we first formulate its normal as follows,
\begin{equation}
    \label{equ:normal_highres}
    n_k=norm(\sum_{j\in N_k}(v_j-v_k)\times (v_{j+1}-v_k))
\end{equation}
where $N_k$ is the set of $v_k$'s neighbors in clockwise order. Then we can formulate the energy term for the shape-from-shading based geometry refinement as follows, 
\begin{equation}
    \label{equ:energy_sfs}
    E_{sfs}=E_{shading}+E_{nearby}+E_{grad}
\end{equation}
where
\begin{equation}
    \label{equ:energy_sfs_shading}
    E_{shading}=\sum_k ||\mathcal{H}_M(\mathcal{L},n_k(\Delta \vec{v}))-\mathcal{I}_{S}(p)||^2
\end{equation}

\begin{figure}
    \centering
    \includegraphics[width=0.9\linewidth]{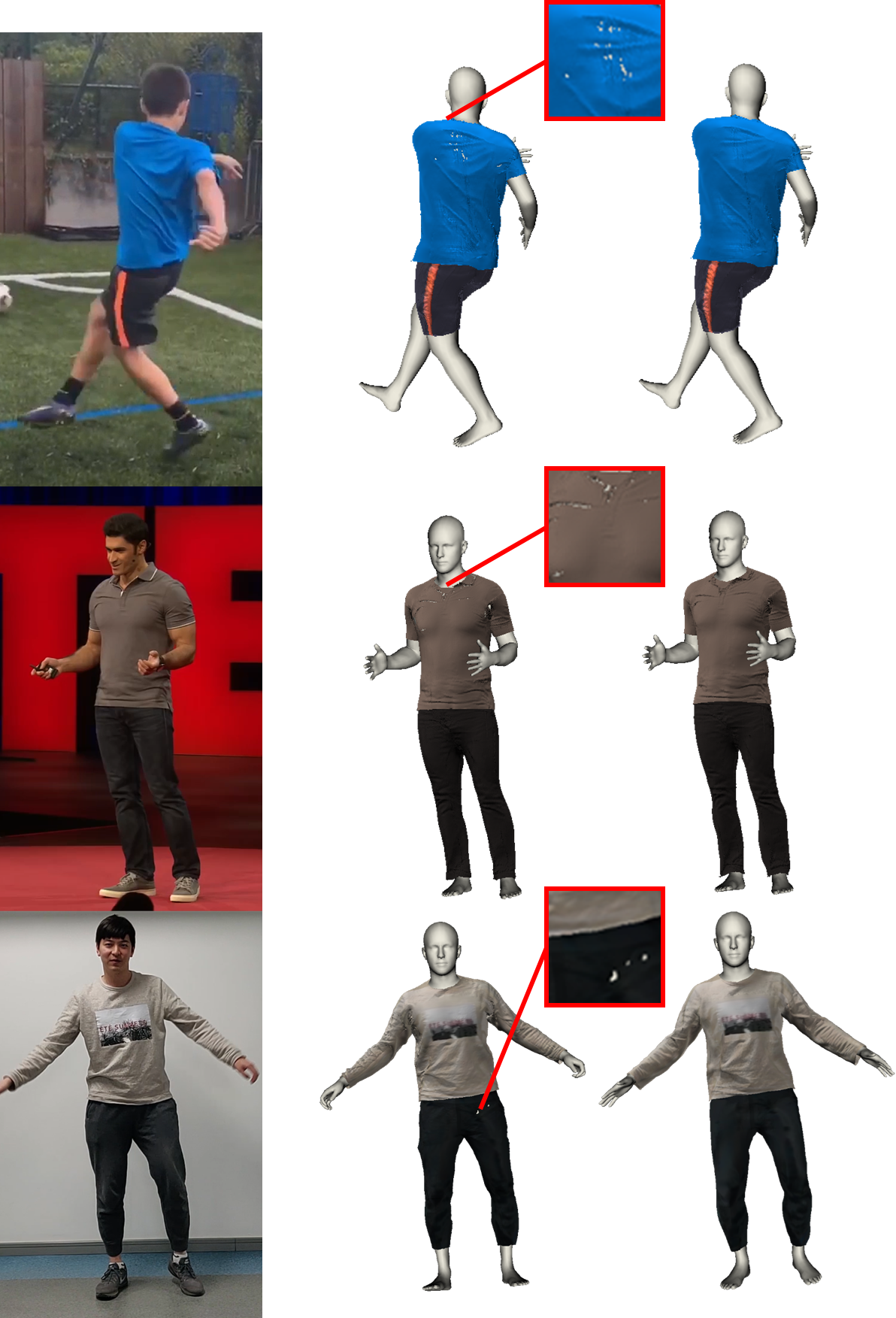}
    \caption{The example results of collision constraint. From left to right: input RGB images, results without collision constraint, results with collision constraint.}
    \label{fig:collision}
\end{figure}

where $\vec{v}$ represents the shifting of all the vertexes, $E_{nearby}$ is the same as in Sect.~\ref{subsec:refinement-non-rigid}, and $E_{grad}$ is for the constraining updating magnitude. Noticing that our reconstructed human contains two layers, i.e., cloth layer and body layer, therefore, traditional shading-based geometry detail enhancement approaches ~\cite{WuShadingHuman} cannot be directly applied, as ~\cite{WuShadingHuman} only generates one-layer mesh and only have to consider mesh self-penetration, while in our pipeline, geometry refinement on the cloth layer may produce penetration or collision with the body layer. As shown in Fig.~\ref{fig:collision}, without collision constraint, there will be some penetrations after refining the geometric details.
To solve this challenge, we propose an iteration optimization approach, by performing shape-from-shading detail enhancement, with physics-based collision detection and resolving in each iteration step. After each physics-based collision resolving, we guarantee penetration-free reconstruction results.

\begin{figure*}
    \centering
    \includegraphics[width=0.9\linewidth]{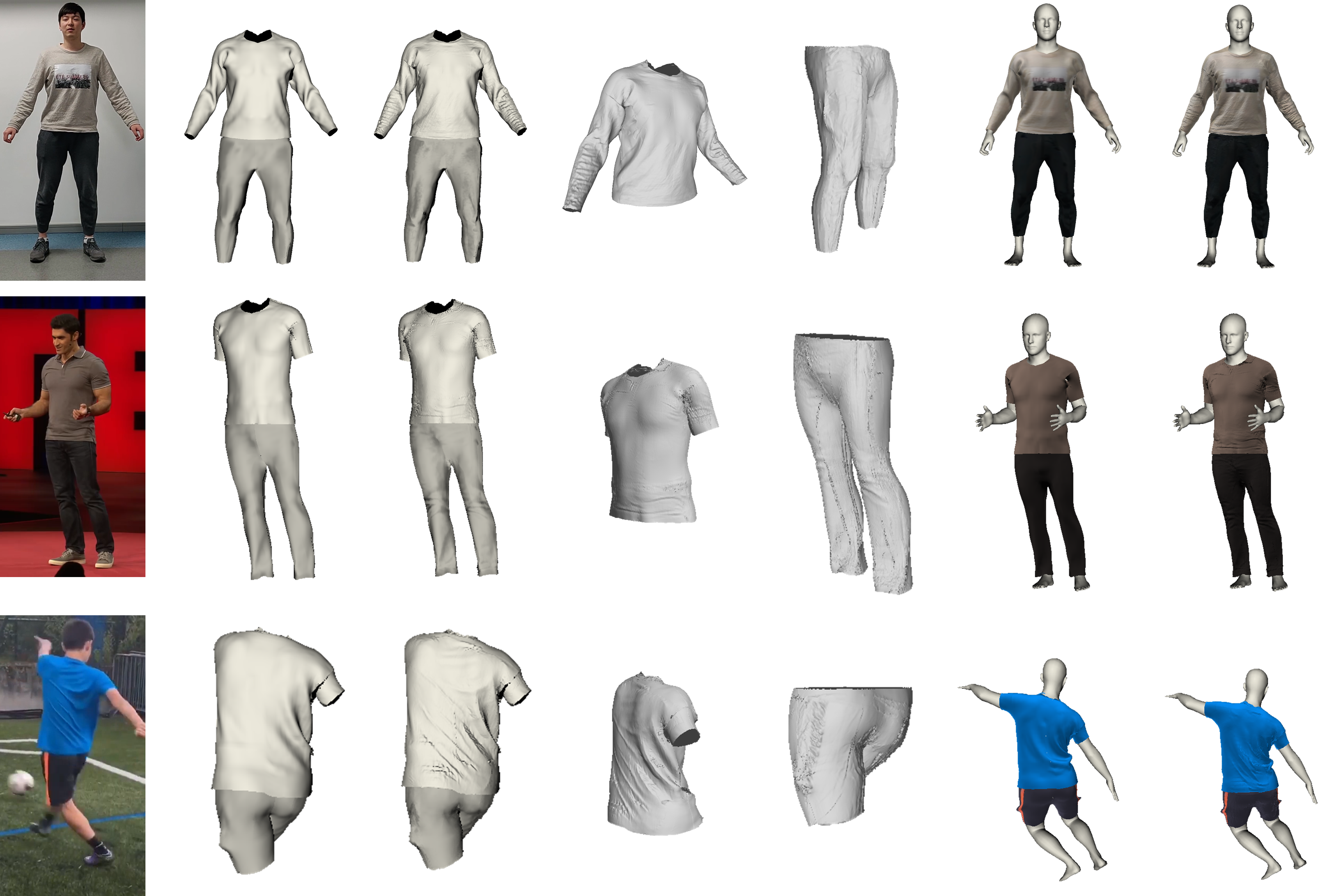}
    \caption{The example result of geometry refinement. From left to right: input RGB image, garment geometry before detail refinement, garment geometry with shape refinement, two garment shapes from another point of view, rendered results without shape refinement procedure, our results.}
    \label{fig:geometry_only}
\end{figure*}

 As shown in Fig.~\ref{fig:geometry_only}, without geometry refinement, the garment geometry lacks details. Meanwhile, after geometry refinement, detail wrinkles and folds can be reconstructed from the input image, making the reconstructed geometry more accurate.

In order to improve the ability of geometry detail representation and time consistency of our dynamic garment reconstruction, we place the vertex displacement in its local coordinate system defined by its normal and neighbor vertexes, which is also used in estimating the displacement of invisible vertices. Note that the motion of each vertex can be decomposed into two parts, namely, the global garment motion by body-garment simulation (see Sect.~\ref{subsec:refinement-non-rigid}) and the local details that cannot be described by simulation. For the invisible vertices, we assume that their local details remain unchanged. Specifically, 
we calculate the global rotation $R$ of the vertex from its value in the visible frame according to the simulated normal orientation, and transform the vertex shifting according to the global rotation to maintain its local-coordinate parameters unchanged. For each temporal frame, we finally add a spatial smooth filtering over the boundaries between the visible and invisible regions, to mitigate the spatially inconsistent seam artifacts.


Finally, given a camera model, as the incident lighting, surface albedo and dynamic geometry details have already been obtained, we render the realistic clothed human performances using spherical harmonics rendering models \cite{basri2001lambertian}. 

\begin{figure*}[ht]
    \centering
    \includegraphics[width=\textwidth]{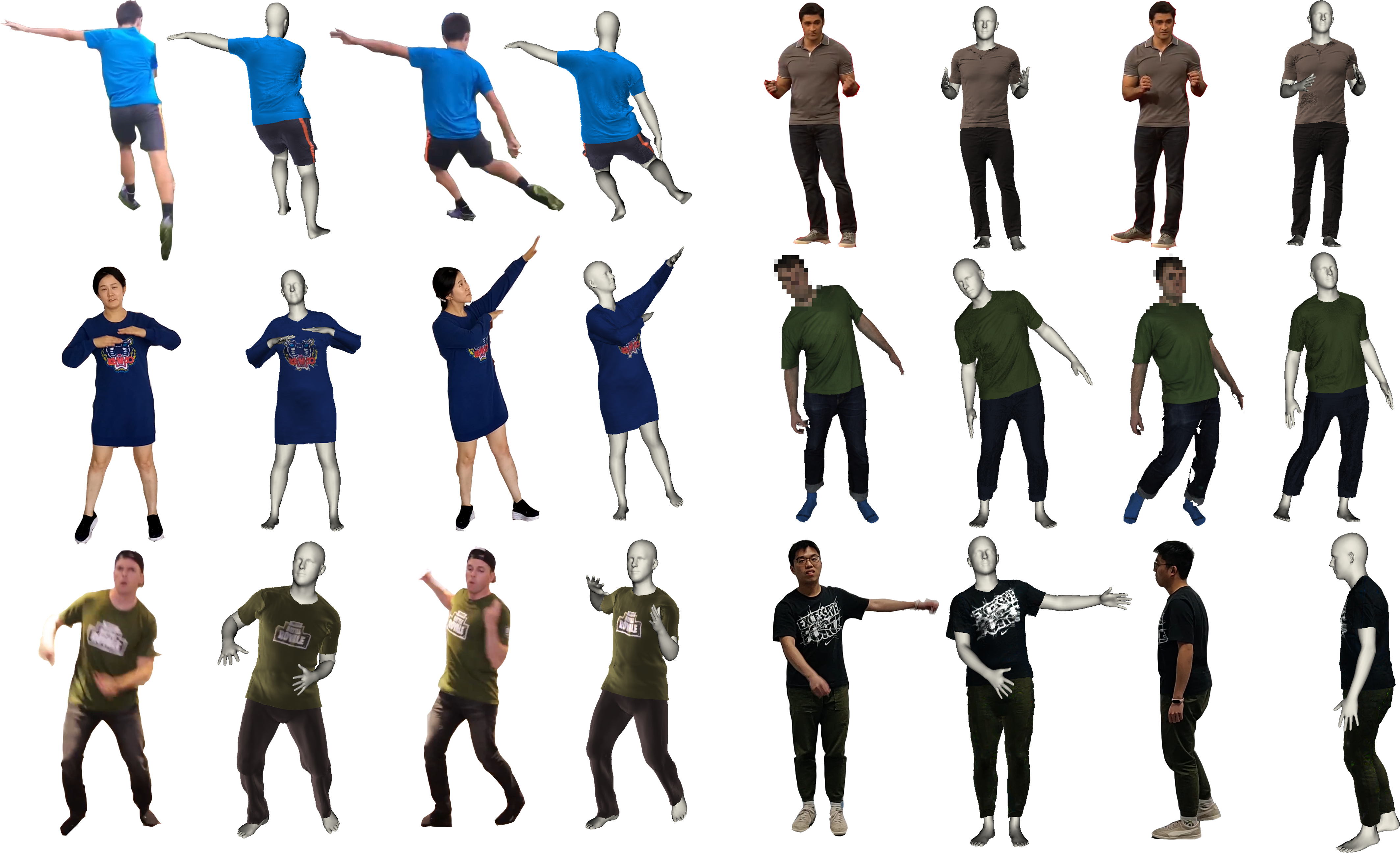}
    \caption{Some reconstruction results in our test sequences. Each pair of result contains the original image on the left and the result on the right.}
    \label{fig:examples}
\end{figure*}

\begin{figure*}[ht]
    \centering
    \includegraphics[width=\textwidth]{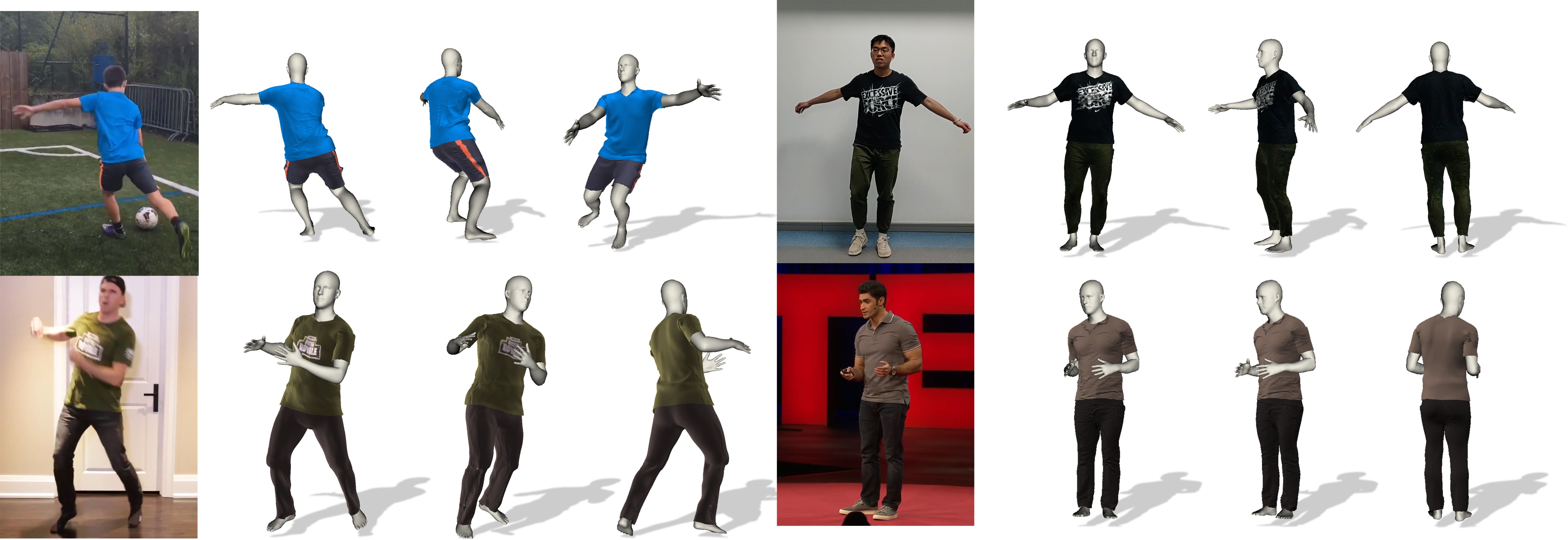}
    \caption{Free-viewpoint rendering of different human models. From left to right: input images, reconstructed models from the captured views, and the reconstructed models in two virtual views.}
    \label{fig:freeview}
\end{figure*}

\section{Experiments}

In our experiments, we use monocular RGB videos from both the internet and our own cameras containing
casual human motions, including walking, playing soccer, speech, exercising, dancing, etc. The human clothing includes pants, trousers, long-sleeve/short-sleeve T-shirt and shirt, which are represented by our designated 2D garment patterns. 

Besides human instance parsing and intrinsic decomposition, the main pipeline takes around 12 hours to process a sequence of 300 frames on a 3.4GHz Intel Xeon E3-1231 processor and an NVIDIA GeForce GTX 1070 GPU. Specifically, the pose and shape estimation takes approximately 15 minutes, garment parameter estimation takes 2 hours for 20 iterations of parameter optimization using every key frame and garment deformation refinement takes 10-12 seconds per frame. After obtaining the deformed mesh, the albedo atlas fusion step takes 10 minutes for cloth albedo generation, and geometry refinement takes 100-120 seconds per frame by using cuSPARSE toolkit. 

\subsection{Qualitative Results}
To evaluate our method, Fig.~\ref{fig:teaser}, Fig.~\ref{fig:examples} and the supplemental video provide the reconstruction results of captured sequences from a monocular video camera, which show that our method is capable of generating plausible human performance capture results with detailed wrinkles and folds, as a benefit of the proposed decomposition-based geometry and albedo refinement method. Note that for the two sequences in the bottom row of Fig.~\ref{fig:examples}, the human characters only perform motions facing to the camera view, without capturing his/her back with turning motions. Nevertheless, our method still generates high quality results for these kinds of motions.

As the albedo map and dynamic geometry details for the cloth mesh are maintained during motion, we can generate free-viewpoint rendering results for the clothed human model.  Fig.~\ref{fig:freeview} shows the 360-degree free-viewpoint rendering of the human, where the cloth details are distinct in different viewpoints. Note that in the second and the last examples of Fig.~\ref{fig:freeview}, the person only shows his front in the whole sequence, but with cloth simulation, we can still render plausible results from the other unseen viewpoints.

\subsection{Comparisons}
\begin{figure*}
    \centering
    \subfigure[]{
    \begin{minipage}[t]{0.48\linewidth}
    \centering
    \includegraphics[width=\linewidth]{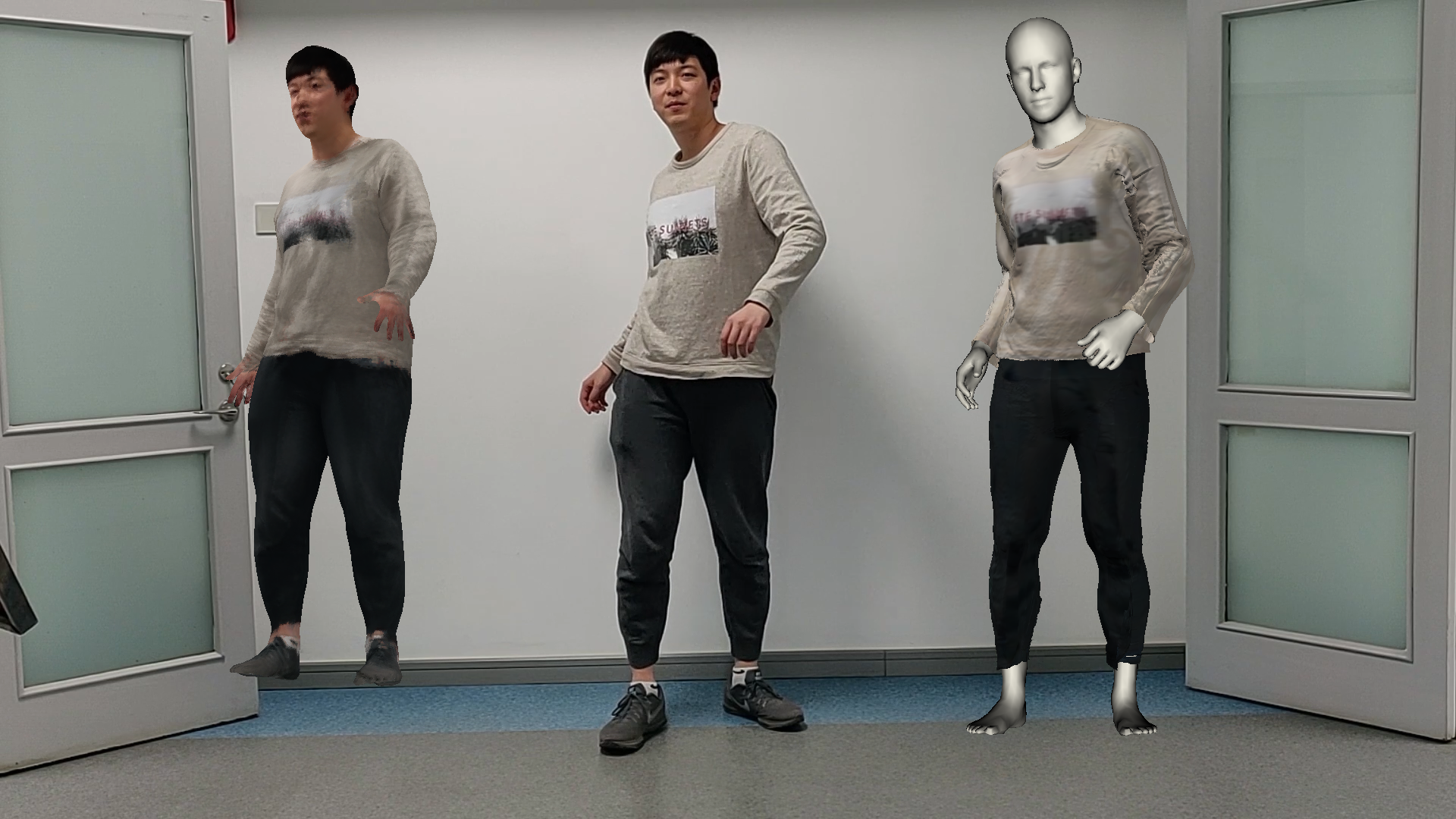}
    \end{minipage}
    }
    \subfigure[]{
    \begin{minipage}[t]{0.48\linewidth}
    \centering
    \includegraphics[width=\linewidth]{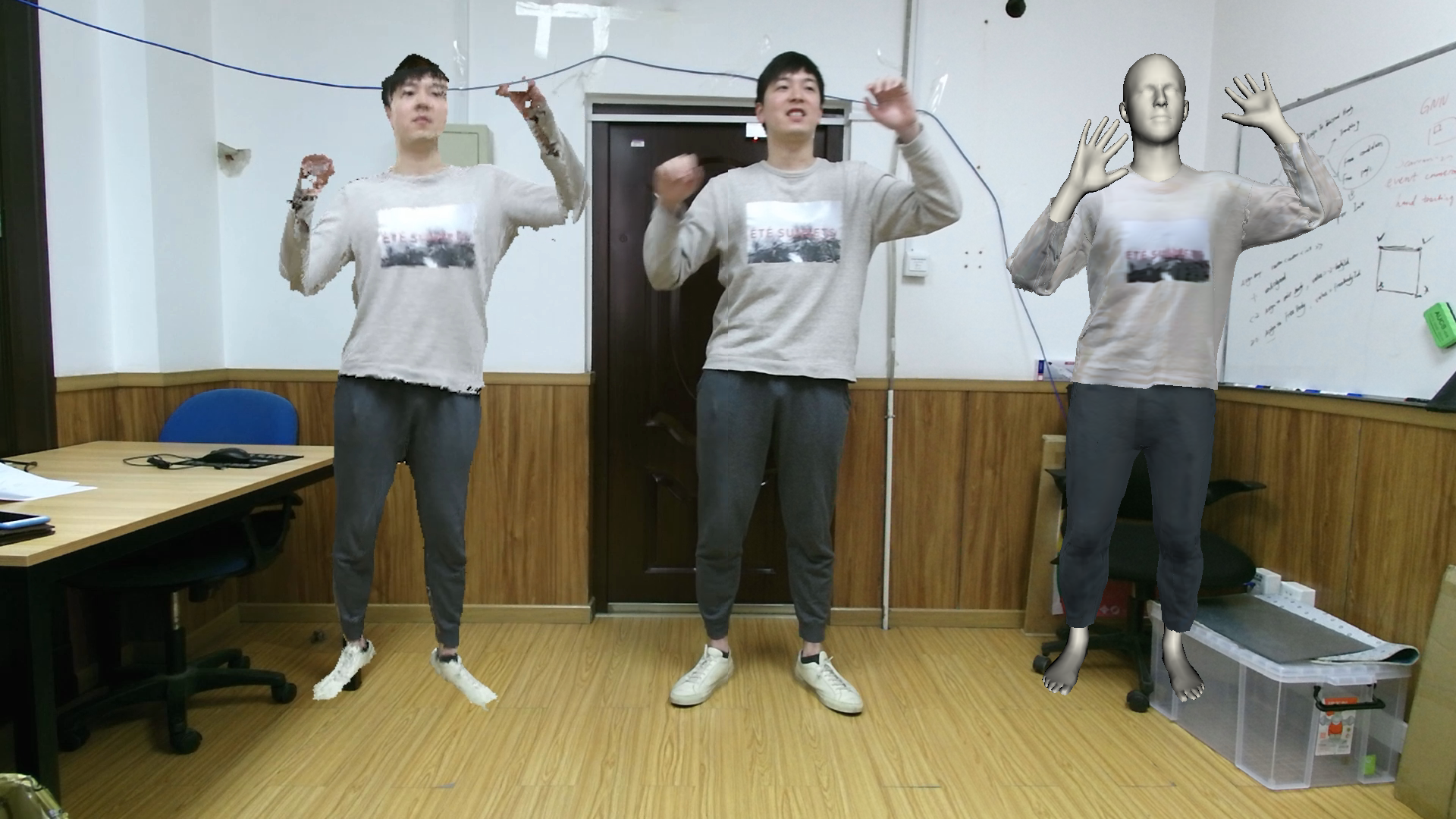}
    \end{minipage}
    }
    \caption{Comparison with \cite{VideoAvater} and \cite{DoubleFusion}. (a) From left to right: \cite{VideoAvater} result, input frame, our result. (b) From left to right: result by a typical non-rigid surface deformation approach using a commercial depth camera \cite{DoubleFusion}, input frame, our result. }
    \label{fig:cmp}
\end{figure*}

\begin{figure}
    \centering
    \includegraphics[width=\linewidth]{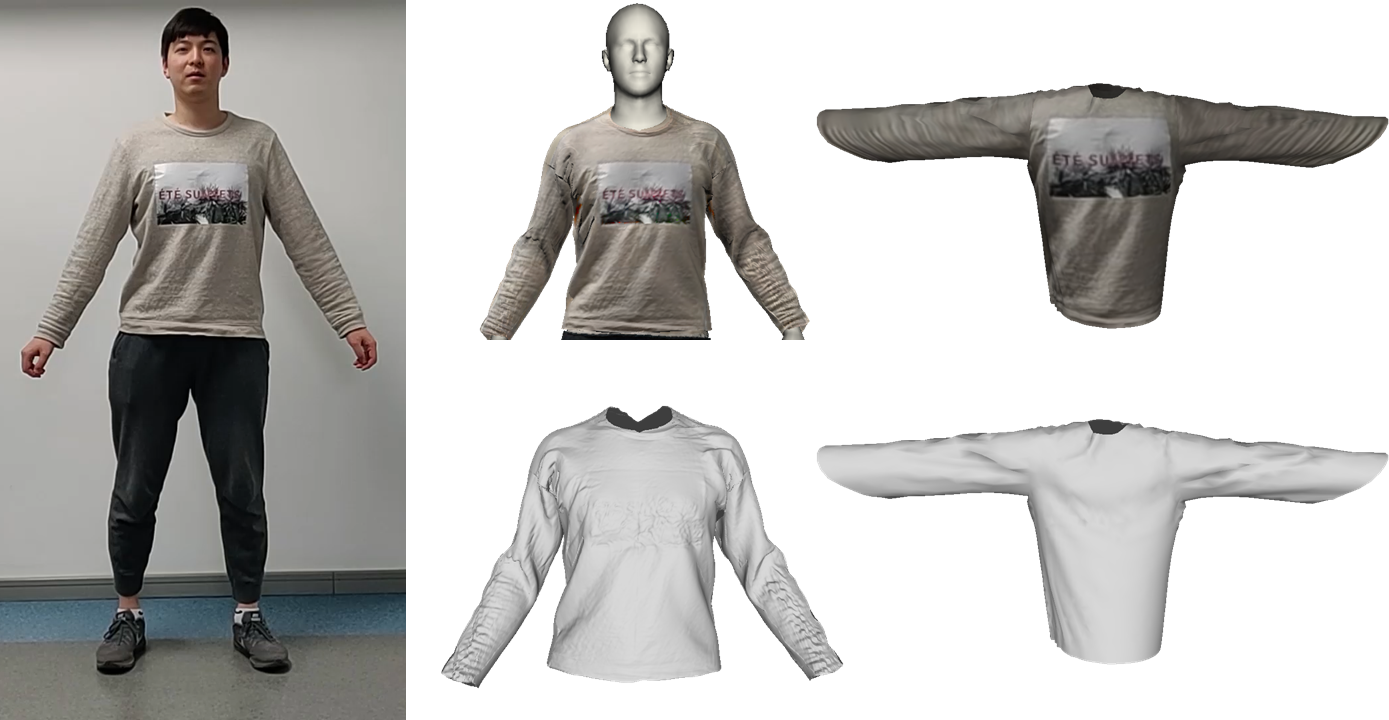}
    \caption{Comparison with ~\cite{vivo19garment}. From left to right: input image, rendered garment and garment geometry generated by our method, rendered garment and garment geometry generated by ~\cite{vivo19garment}. Notice that ~\cite{vivo19garment} generates T-pose clothing output.}
    \label{fig:cmp_vivo}
\end{figure}

\begin{figure*}
    \centering
    \subfigure[]{
    \begin{minipage}[t]{0.43\linewidth}
    \centering
    \includegraphics[width=\linewidth]{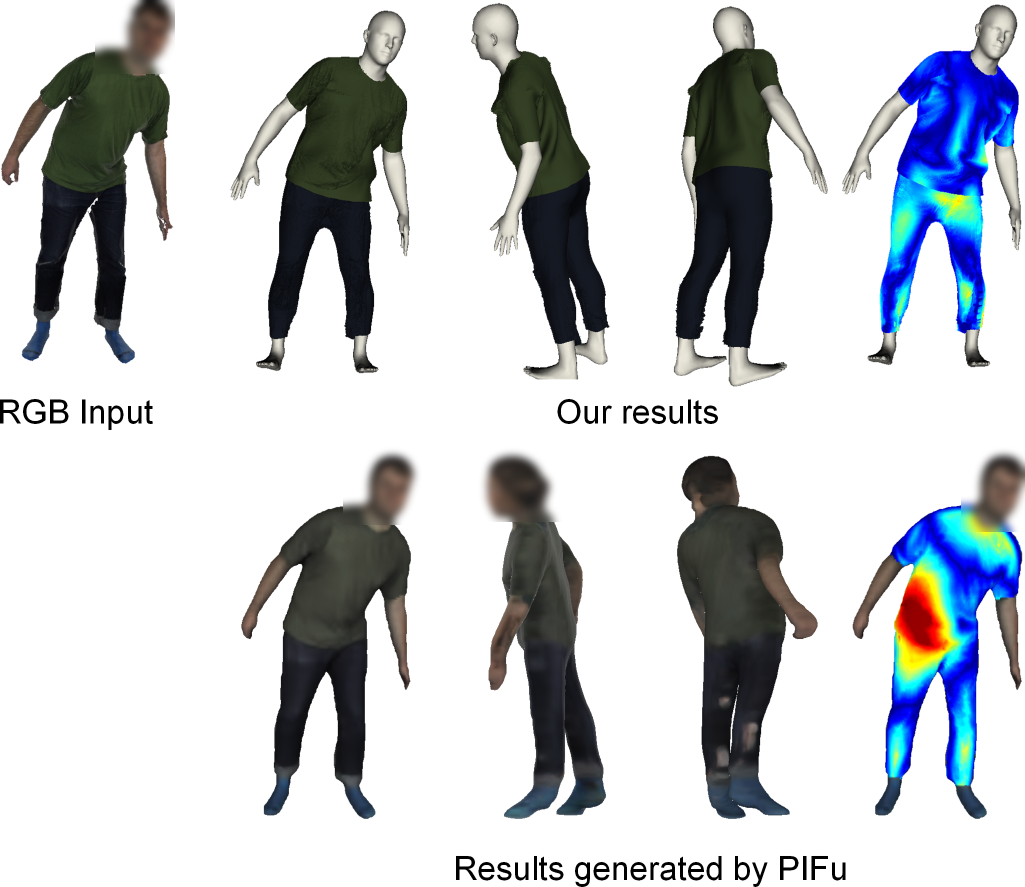}
    \end{minipage}
    }
    \subfigure[]{
    \begin{minipage}[t]{0.53\linewidth}
    \centering
    \includegraphics[width=\linewidth]{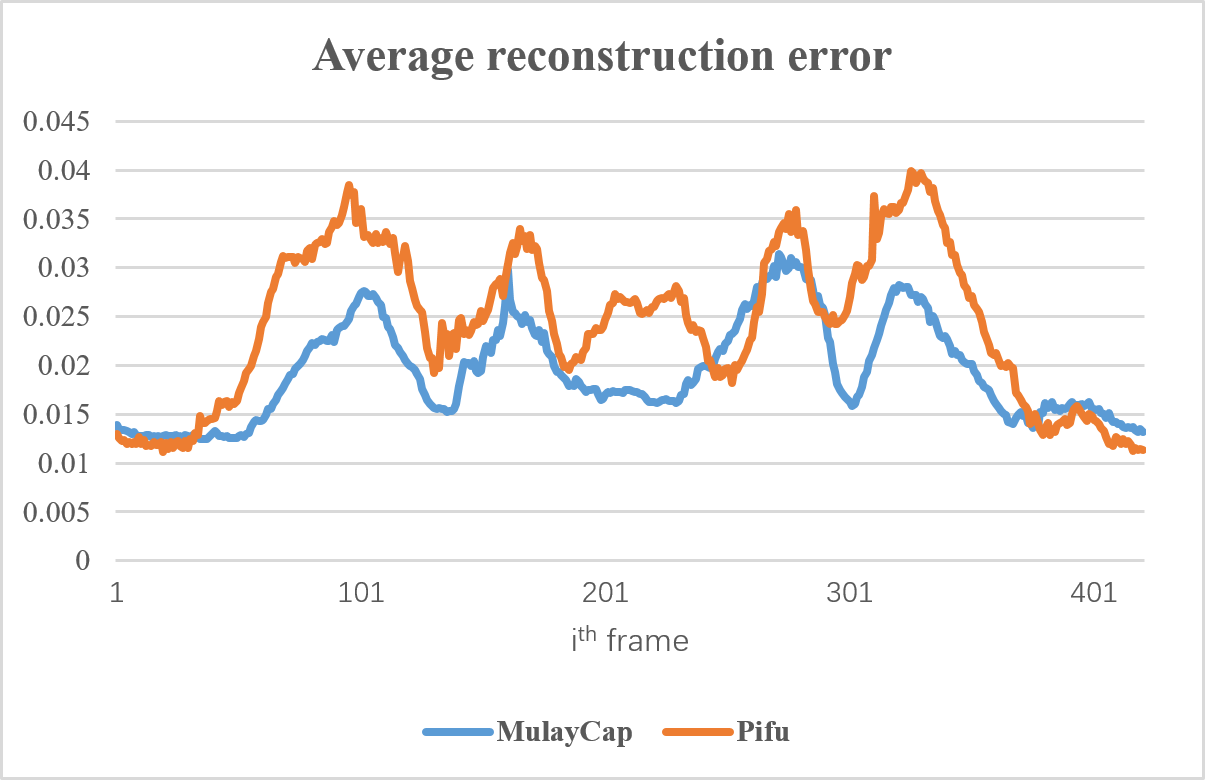}
    \end{minipage}
    }
    \caption{Qualitative and quantitative comparison with PIFu \cite{pifuSHNMKL19} using rendered 4D model in BUFF Dataset\cite{Zhang_2017_CVPR} as input. (a) From left to right: rendered human model, reconstruction results from different viewpoints by MulayCap and PIFu\cite{pifuSHNMKL19}, error map. (b) Quantitative comparison between two methods in one 4D sequential using per-vertex average error. }
        \vspace{-12pt}
    \label{fig:eval_pifu}
\end{figure*}

We compared our human performance capture results with \cite{VideoAvater} and typical template-based deformation methods \cite{li2009robust,guo2015robust} using a commercial RGBD camera, as shown in Fig.~\ref{fig:cmp} and the supplemental video. The video avatar reconstruction method in \cite{VideoAvater} takes a single view video of human performance as input, and rectifies all the poses in the image frames to a T-pose for bundle optimization of shape. However, the subject needs to perform the restrictive movement to allow accurate shape reconstruction. So it fails to work for other more generate shapes, poses and dynamic textures, as shown in Fig.~\ref{fig:cmp}(a). In contrast,  our method works robustly even when subjects perform more casual motions with natural cloth-body interaction and dynamic texture details.   

Fig.~\ref{fig:cmp}(b) shows the comparison with typical template-based deformation approach \cite{li2009robust,guo2015robust}. The result on the left is obtained by first fusing the geometry and texture using the DoubleFusion \cite{DoubleFusion} system, followed by skeleton driven non-rigid surface deformation to align with the depth data and the silhouette. As shown in Fig.~\ref{fig:cmp}(b) and the video, the texture of such non-rigid reconstruction is static, so it cannot dynamically model changing surface details. In contrast, our method is able to capture the dynamical winkles and produce more plausible garment deformations. 

We also make a comparison with a model-based approach ~\cite{vivo19garment} on our data. As their method takes only one picture as input, we also take only one picture and feed it into our pipeline for a fair comparison. 
Notice that ~\cite{vivo19garment} generates T-pose garment mesh only. 
As shown in Fig.~\ref{fig:cmp_vivo}, regarding to the garment geometry, ~\cite{vivo19garment} generates an over-smoothed surface of the garment without detailed wrinkles and folds, and the shape of the cloth does not fit the input image accurately. Meanwhile, our method successfully recovers the geometric details and produce much more realistic rendering results.

We also make a quantitative evaluation on BUFF Dataset~\cite{Zhang_2017_CVPR} and compare MulayCap quantitatively with PIFu~\cite{pifuSHNMKL19}, which is deep learning method for reconstructing clothed human body from a single image, also without a pre-scanned template. The reconstruction results and the per-vertex average error is shown in Fig.~\ref{fig:eval_pifu}. As shown in Fig.~\ref{fig:eval_pifu}(a), benefiting from our multi-layer representation of the model and physics-based cloth simulation, we can generate high-frequency details of the cloth, both on the front and back. The pose estimated is also consistent with the input image. Meanwhile, although the model generated by PIFu~\cite{pifuSHNMKL19} looks plausible from the front view, we can see that it actually generates a wrong pose of the human, also the texture on the back is not so vivid neither realistic. The comparison with PIFu~\cite{pifuSHNMKL19} can be regarded as a typical comparison between the model-based methods and data-driven generative methods. Benefiting from other model-based methods like HMMR~\cite{3DHumanDynamics}, we can generate more robust and accurate garment results. On the contrary, the implicit representation of PIFu~\cite{pifuSHNMKL19} limits its ability of using model-based priors, leading to unrealistic human pose and texture generated. 

As for quantitative experiments, we first put the model from both PIFu~\cite{pifuSHNMKL19} and MulayCap into a consistent coordinate with the ground truth 3D model of BUFF Dataset~\cite{Zhang_2017_CVPR}, and then align the models with the ground truth one using ICP for solving the scale and relative transition of the models. The error is evaluated using the nearest-neighbor L2 distance. Fig.~\ref{fig:eval_pifu}(b) shows that the per-vertex error of PIFu~\cite{pifuSHNMKL19} is larger than MulayCap in most frames of an input video sequence rendered from BUFF Dataset~\cite{Zhang_2017_CVPR}, which shows that with our multi-layer human performance capture method, we can generate more accurate results than the one generated using an end-to-end network.

\subsection{Applications}
With our proposed multiple-layer modeling for human performance capture, our method produces fully-semantic reconstruction and enables abundant editing possibilities in the following applications.

\begin{figure}
    \centering
    \includegraphics[width=0.38
    \textwidth]{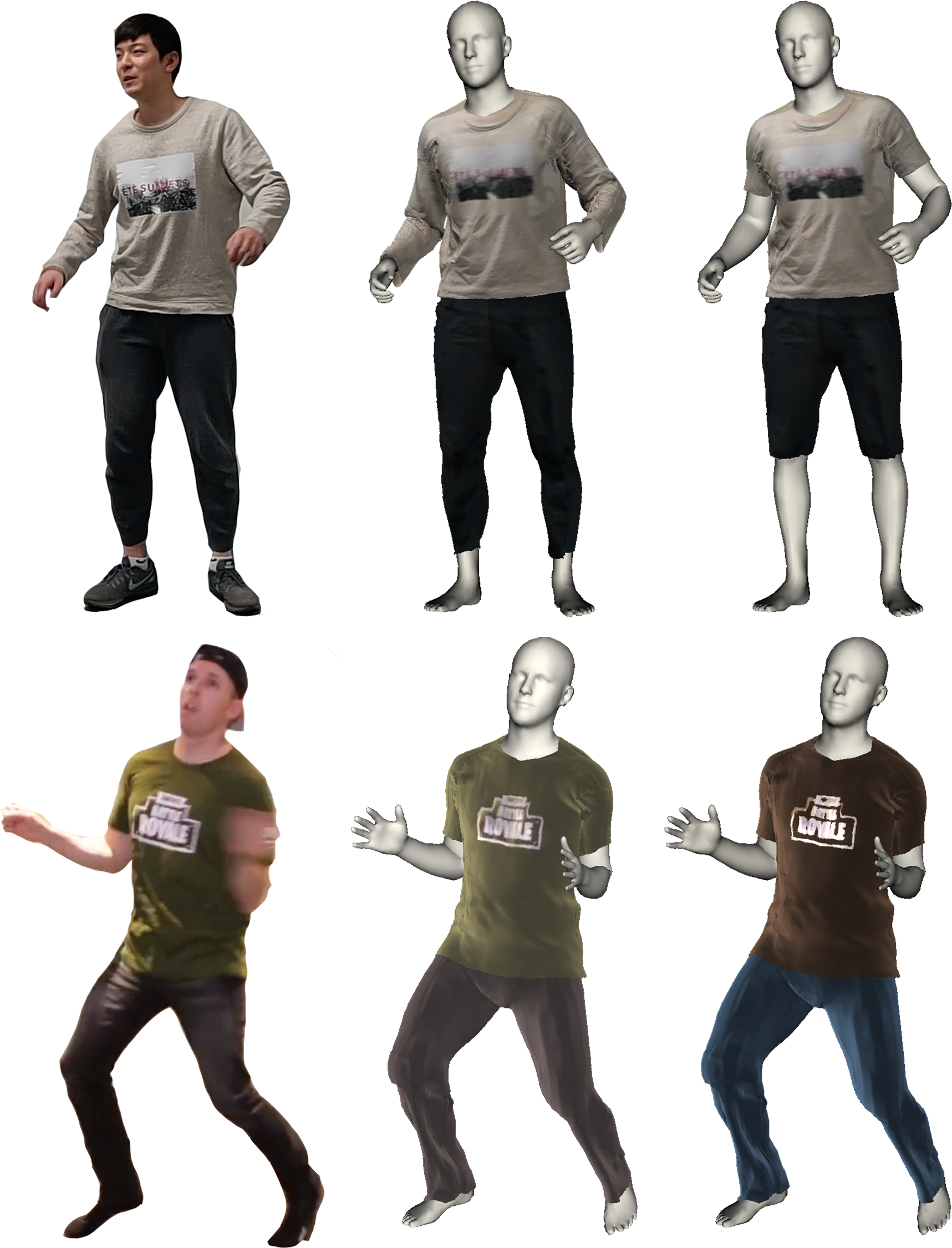}
    \caption{Garment editing results. The upper row is garment shape editing and the bottom row is for garment color editing. From left to right: input RGB frames, reconstructed results, results with shape editing and color editing.}
    \vspace{-12pt}
    \label{fig:cloth_editing}
\end{figure}

\textit{Garment Editing.} Since in our method semantically models garments on the shape and texture, garment editing in terms of shape or texture can be achieved, as demonstrated in Fig.~\ref{fig:cloth_editing}. Garment shape editing (upper row) allows the change of the length parameters of the T-shirt sleeve and the trousers so that the human performance of the same character with new clothing can be obtained. By combining a new albedo color of the cloth with the original shading results, we can render realistic color editing results for the reconstructed human performance as shown in the bottom row.

\vfill

\begin{figure}
    \centering
    \includegraphics[width=0.45\textwidth]{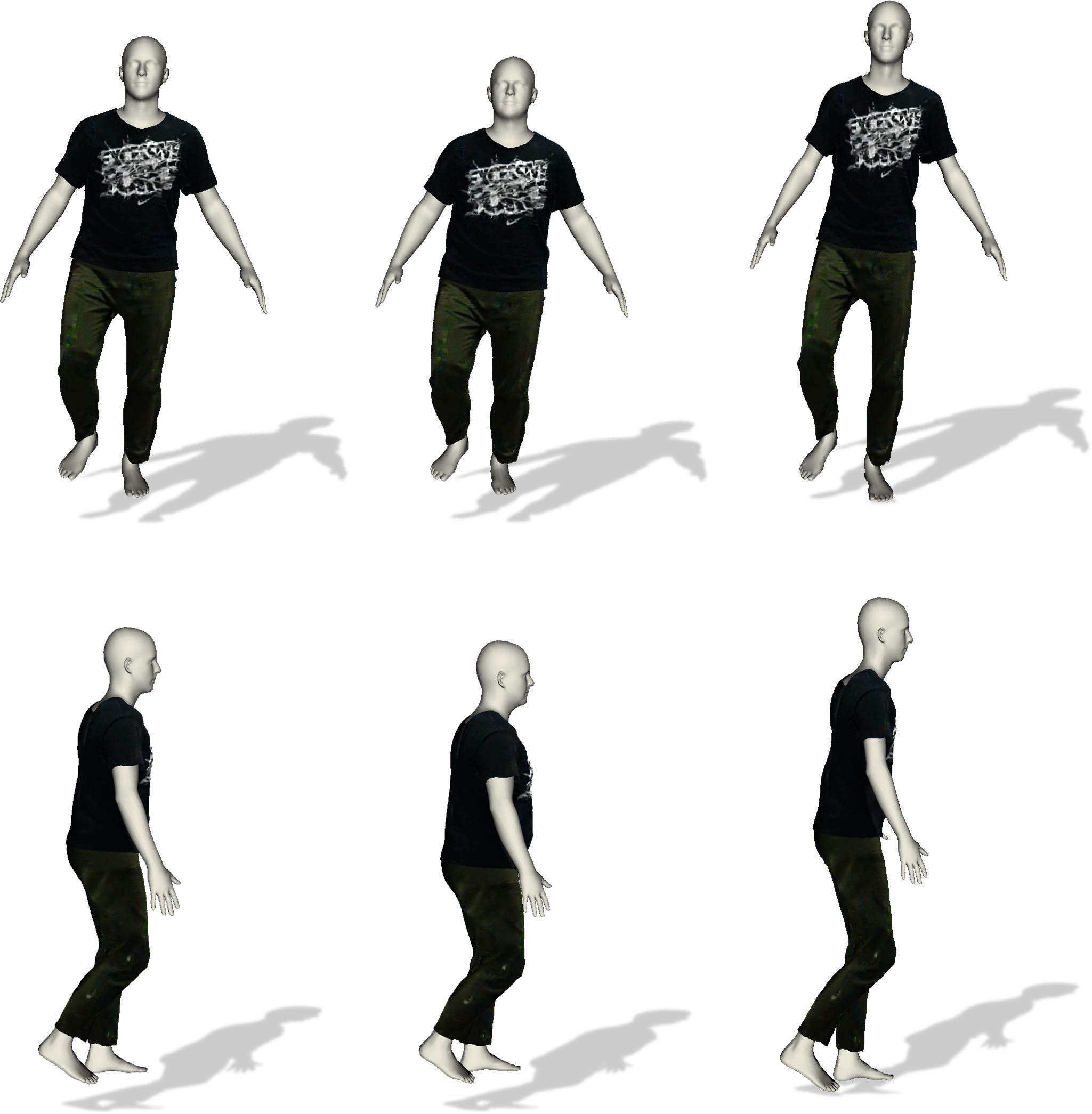}
    \caption{Clothing retargeting between different human bodies. From left to right: reconstructed clothed human models, and two retargeting results using a taller thin body shape and a fatter body shape.}
    \label{fig:retargeting}
\end{figure}

\textit{Retargeting.} After the cloth shape and albedo have been generated for a sequence, we can retarget the clothing to other human bodies. Recall that the human models represented by SMPL model, which guarantees topology-consistency between different human models. So we can calculate a non-rigid warp field between the two human bodies with different shapes but the same pose, and adopt this warp field for cloth vertex mapping between the two models. The result is shown in Fig.~\ref{fig:retargeting}, where two target body shapes are used for the retargeting application.

\begin{figure}[ht]
    \centering
    \subfigure{
    \begin{minipage}[t]{0.45\linewidth}
    \centering
    \includegraphics[width=\linewidth]{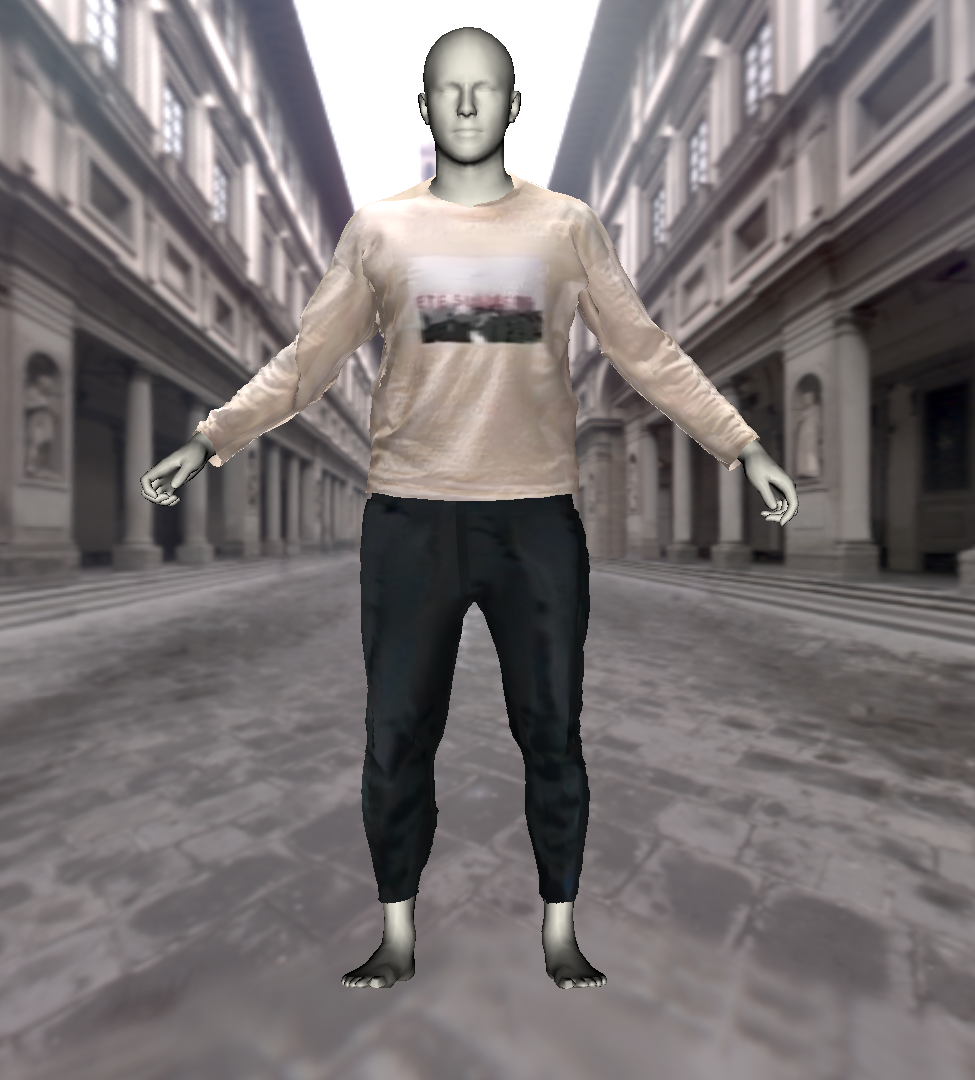}
    \end{minipage}
    }
    \subfigure{
    \begin{minipage}[t]{0.45\linewidth}
    \centering
    \includegraphics[width=\linewidth]{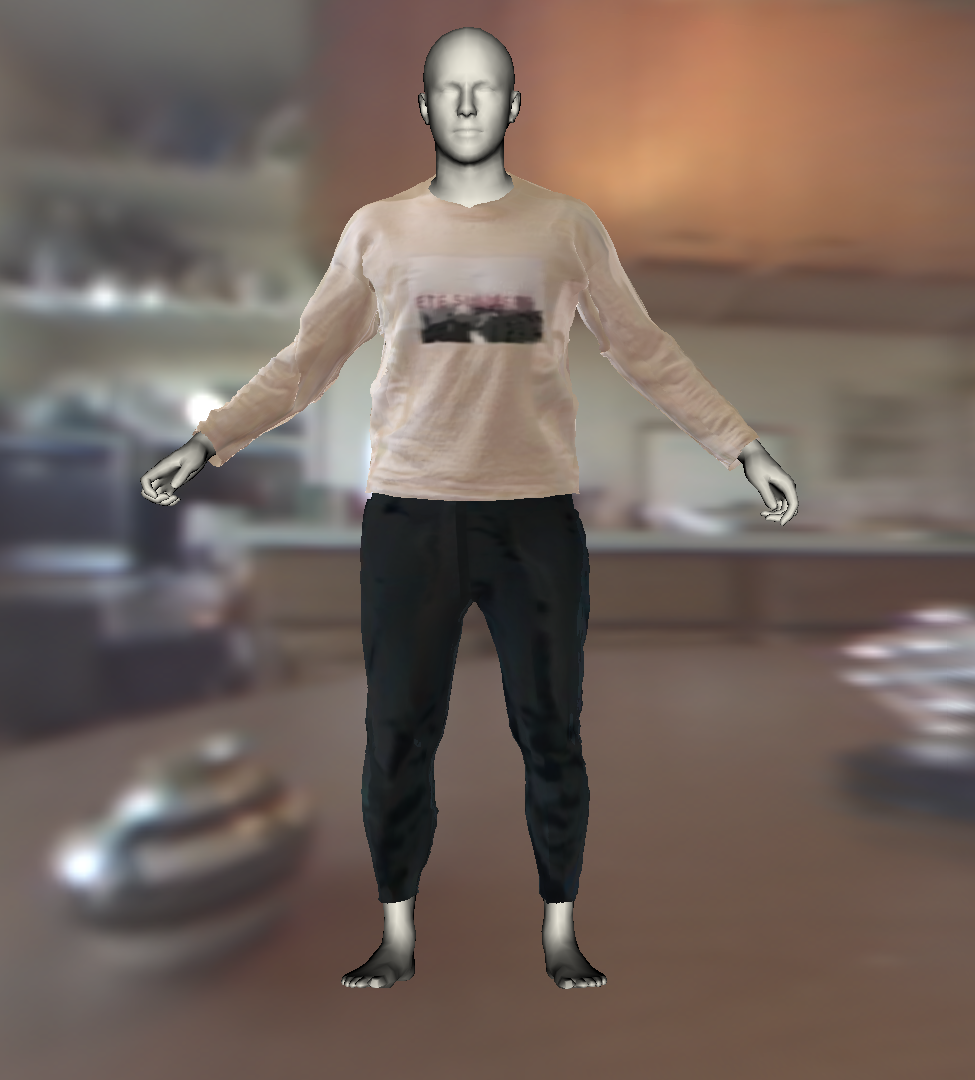}
    \end{minipage}
    }
    \subfigure{
    \begin{minipage}[t]{0.45\linewidth}
    \centering
    \includegraphics[width=\linewidth]{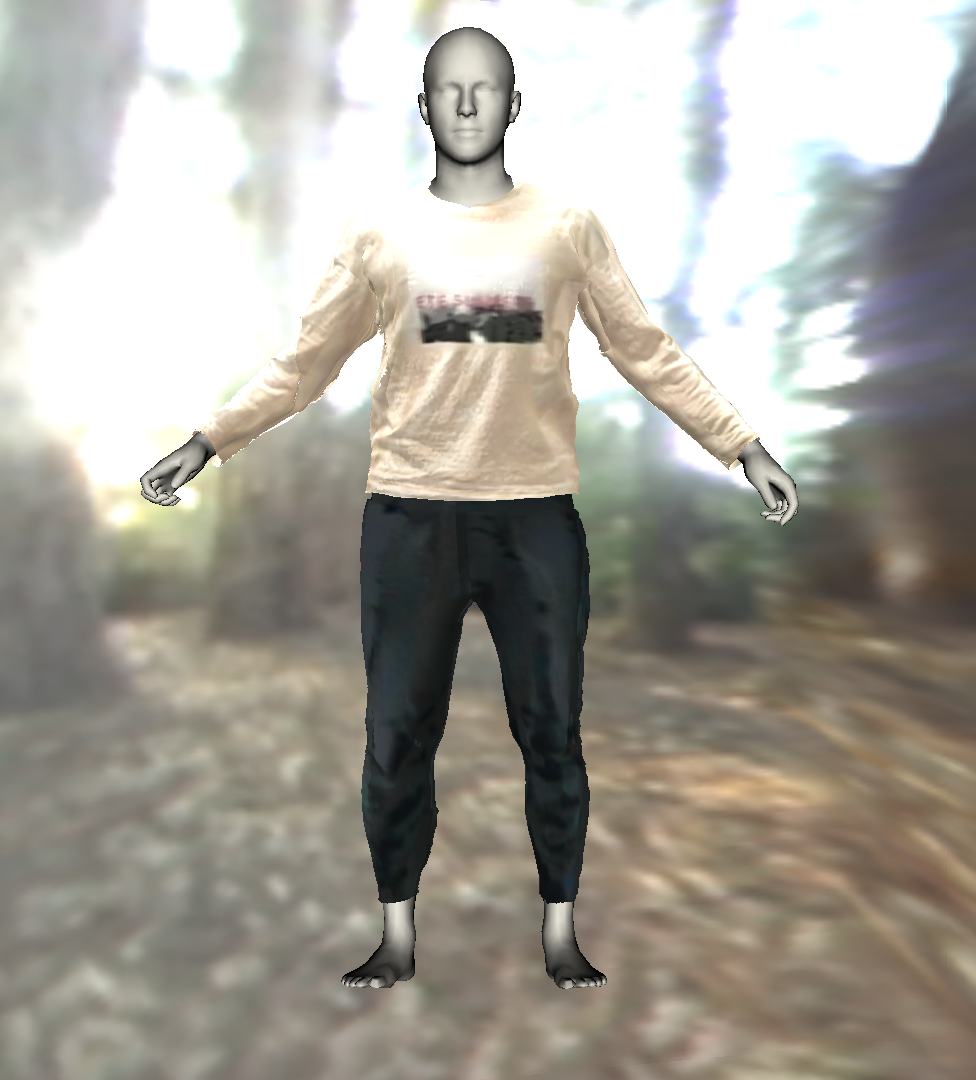}
    \end{minipage}
    }
    \subfigure{
    \begin{minipage}[t]{0.45\linewidth}
    \centering
    \includegraphics[width=\linewidth]{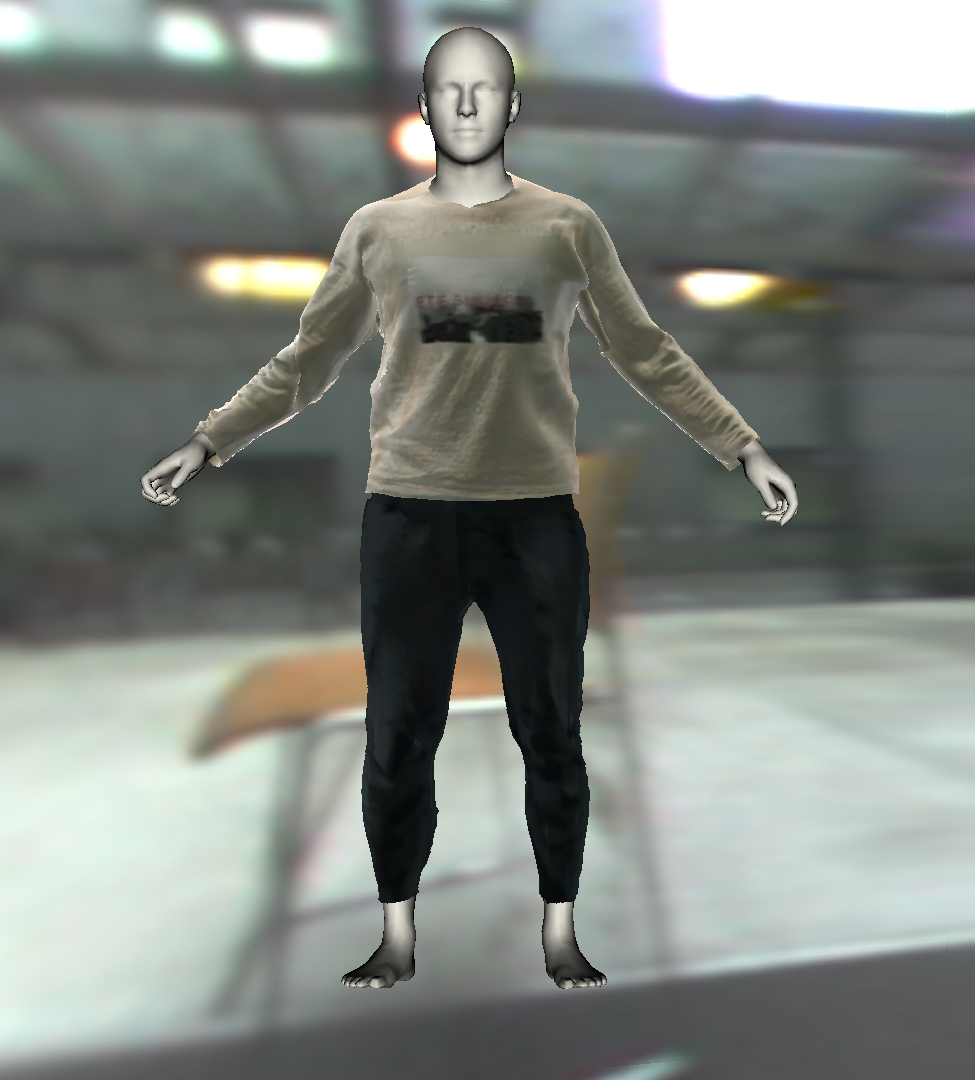}
    \end{minipage}
    }
    \caption{Relighting results in four different environmental illumination maps from ~\cite{Debevec98renderingsynthetic}.}
    \label{fig:relighting}
\end{figure}

\textit{Relighting.} Given albedo and detailed geometry with wrinkles and folds of the garment, we can generate relighting results for the captured sequence. As shown in Fig.~\ref{fig:relighting}, we put the character in four different environment illuminations and apply the relighting using spherical harmonic lighting coefficients generated by the cube-map texture. The albedo and geometry details are consistent in different lighting environments.

\begin{figure}
    \centering
    \subfigure{
    \begin{minipage}[t]{0.45\linewidth}
    \centering
    \includegraphics[width=\linewidth]{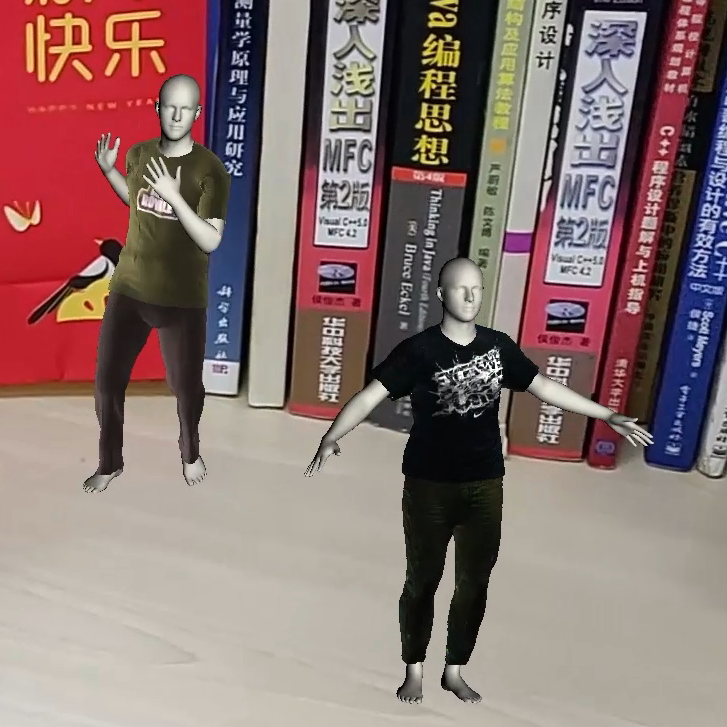}
    \end{minipage}
    }
    \subfigure{
    \begin{minipage}[t]{0.45\linewidth}
    \centering
    \includegraphics[width=\linewidth]{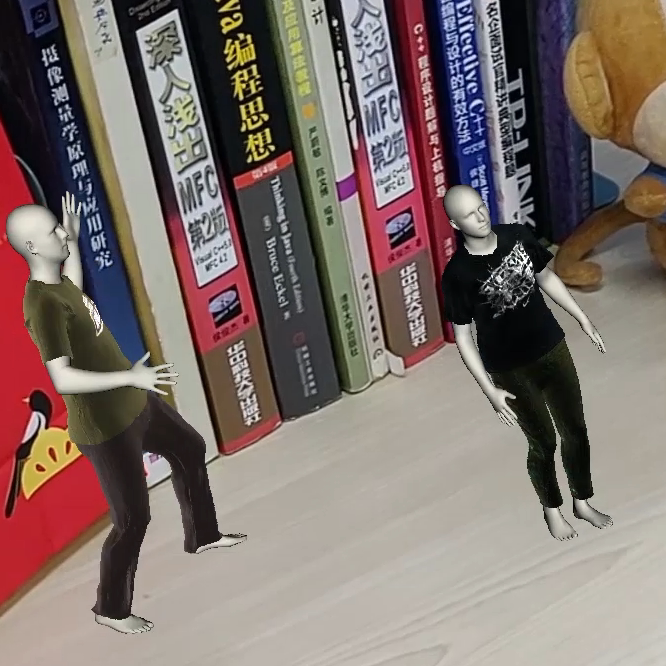}
    \end{minipage}
    }
    \caption{Two frames of an augmented-reality application. We estimate the camera parameters using \cite{MVE} and render the clothed human performance on the desk.}
    \label{fig:desk}
\end{figure}

\textit{Augmented Reality.} As we can automatically generate 4D human performance with only RGB video, it can be integrated into a real video for VR/AR applications. Given a video sequence of a particular scene as well as the camera positions and orientations in each frame, we can render the human performance at a particular location in the scene. With AR glasses such as Hololens, observers can see human performance in any viewpoint. The examples of such mixed-reality rendering are shown in Fig.~\ref{fig:desk} and the supplemental video.

\begin{figure}
    \centering
    \subfigure[]{
    \begin{minipage}[t]{0.45\linewidth}
    \centering
    \includegraphics[width=\linewidth]{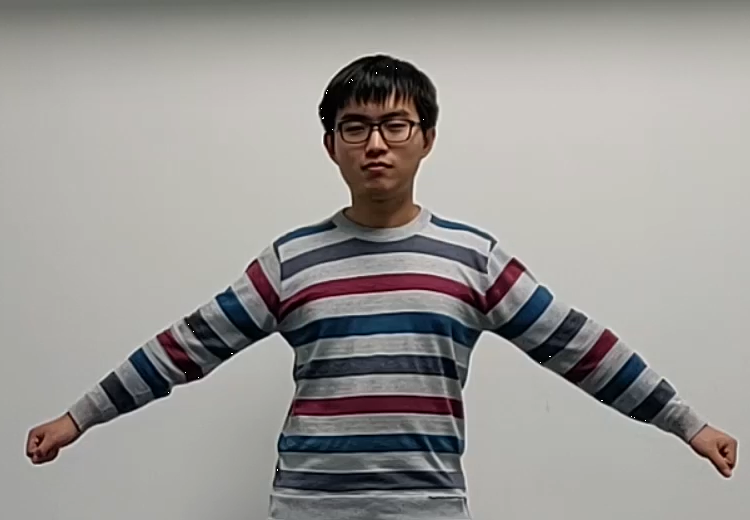}
    \end{minipage}
    }
    \subfigure[]{
    \begin{minipage}[t]{0.45\linewidth}
    \centering
    \includegraphics[width=\linewidth]{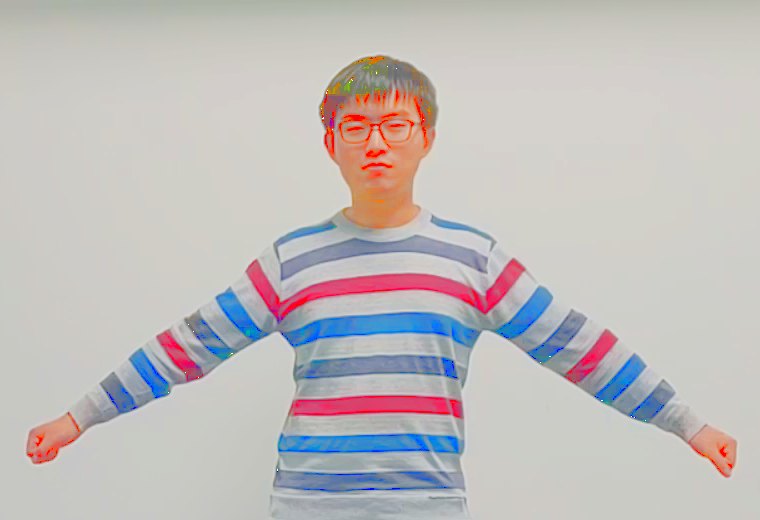}
    \end{minipage}
    }
    \subfigure[]{
    \begin{minipage}[t]{0.45\linewidth}
    \centering
    \includegraphics[width=\linewidth]{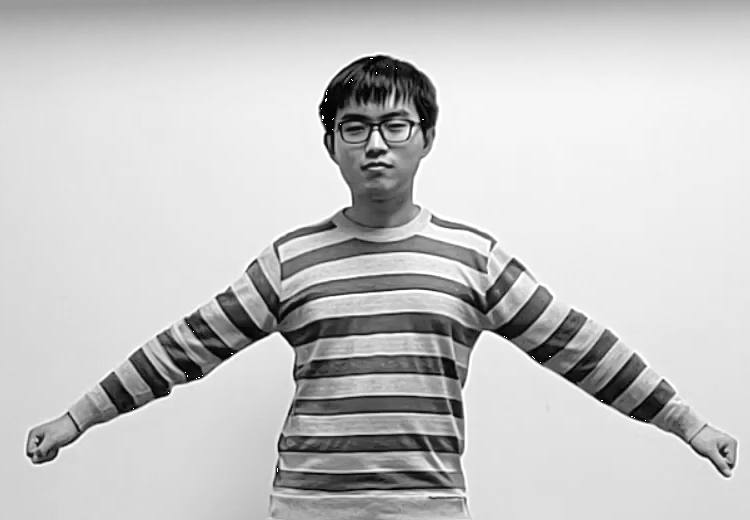}
    \end{minipage}
    }
    \subfigure[]{
    \begin{minipage}[t]{0.45\linewidth}
    \centering
    \includegraphics[width=\linewidth]{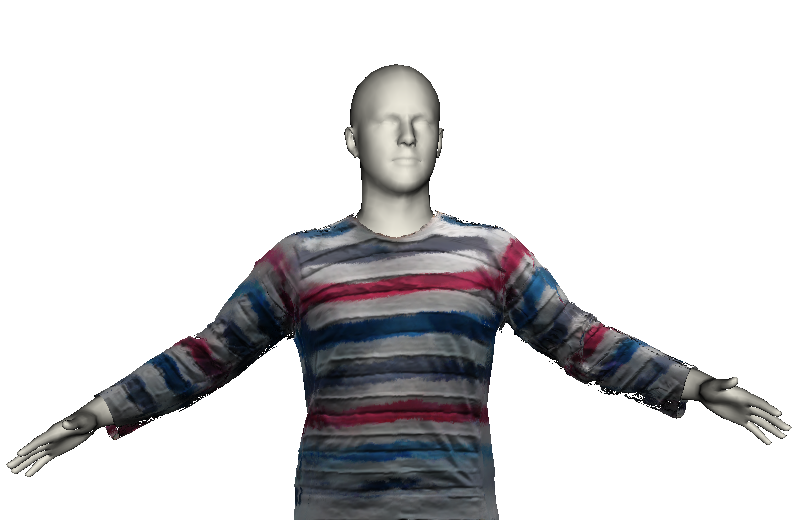}
    \end{minipage}
    }
    \caption{Illustration of the failure case. (a) The input image. (b) The decomposed albedo image. (c) The decomposed shading image. (d) The rendered result.}
    \label{fig:limitation}
\end{figure}

\section{Conclusion}
In this paper, we present a novel method, called {\em MulayCap}, based on a multi-layer decomposition of geometry and texture for human performance capture using a single RGB video. Our method can generate novel free-view rendering of vivid cloth details and human motions from a casually captured video, either from the internet or video captured by the user. There are three main advantages of MulayCap: (1) it obviates the need for tedious human specific template scanning before real performance capture and still achieved high quality geometry reconstruction on the clothed human performances. This is made possible through the proposed GfV method based on cloth simulation techniques for estimating garment shape parameters by  fitting the garment appearance to the input video sequence; (2) MulayCap achieves realistic rendering of the dynamically changing details on the garments by using a novel technique of decoupling texture into albedo and shading layers. It is worth noting that such dynamically changing textures have not been demonstrated in any existing monocular human performance capture systems before; finally (3) benefiting from the fully semantic modeling in MulayCap, the reconstructed 4D performance naturally enables various important editing applications, such as cloth editing, re-targeting, relighting, etc. 


\textbf{Limitation and Discussion:} 
MulayCap mainly focuses on the body and garment reconstruction, while the other semantic elements like head, facial expression, hand, skin and shoes would require extra efforts to be handled properly. Another deficiency is that the body motion still suffers from jittering effects, as the body shape parameters are difficult to be accurately {\em and} smoothly estimated from the video based on the available human shape and pose detection algorithms \cite{humanMotionKanazawa19}. As a consequence, we cannot handle fast and extremely challenging motions, as the pose detection on challenging motions contains too many errors for cloth simulation and garment optimization. Also, although our system is robust for common cases, our cloth pattern cannot handle all possible clothes or clothes with non-common shapes.

In addition, in our pipeline, the qualities of the albedo and shading image are crucial for the final rendering results, which may be affected by the performance of intrinsic decomposition methods to a certain extent. For garments with complex texture patterns such as the lattice T-shirt shown in Fig.~\ref{fig:limitation}, existing intrinsic decomposition methods can hardly produce accurate results. In our case, since the shading image extracted by \cite{nestmeyer2017reflectance} still contains much albedo information, the geometry detail solved by our system is messed with albedo information, as shown in Fig.~\ref{fig:limitation}. As most of the existing intrinsic decomposition methods are intended for general scenes, a novel intrinsic decomposition method particularly designed for garments may further improve the shading and albedo estimation in our task.

As for the future work, a more precise human performance capture including hands, skins, shoes, etc., as well as a variety of garment patterns like skirts, coats, etc. are promising directions to be explored. Along with the booming of single image human body estimation research \cite{MonocularTotalCap,3DHumanDynamics}, research attentions can be directed on how to achieve jittering-free motion reconstruction to handle more challenging motions. Overall, we believe that our paper may inspire much follow-up research towards improving the quality of convenient and efficient human performance capture using a single monocular video camera, thus facilitating and promoting applications of consumer level human performance capture.

\vspace{-10pt}
\ifCLASSOPTIONcompsoc
  \section*{Acknowledgments}
\else
  \section*{Acknowledgment}
\fi

The authors would like to thank Tsinghua University and The Hong Kong University of Science and Technology for supporting this work.

\ifCLASSOPTIONcaptionsoff
  \newpage
\fi
\vspace{-10pt}



\bibliographystyle{IEEEtran}
\bibliography{bibliography}
%



%
\vspace{-20pt}

\begin{IEEEbiography}[{\includegraphics[width=1in,height=1.25in,clip,keepaspectratio]{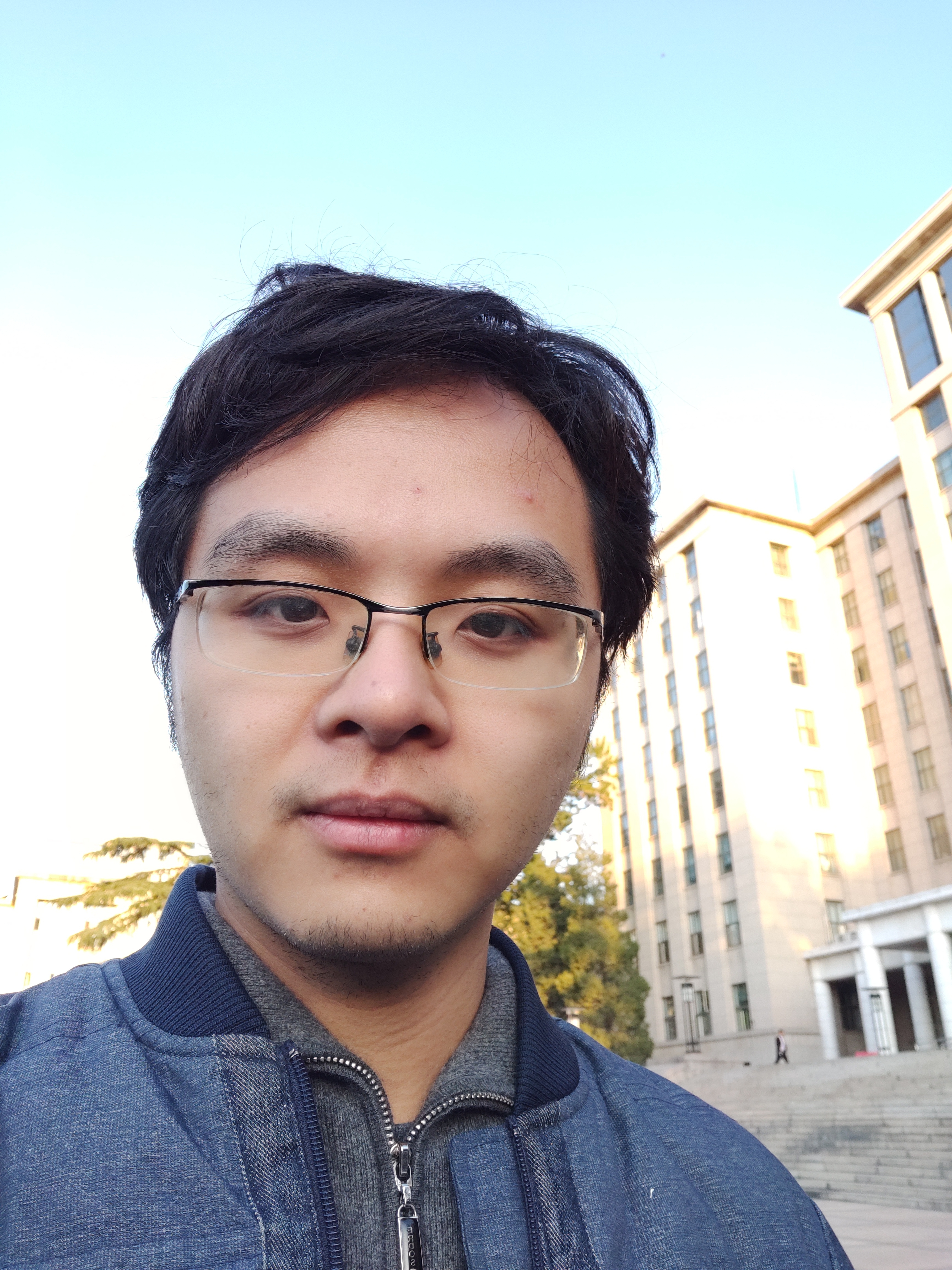}}]{Zhaoqi Su}
received the B.S. degree in Department
of Physics, Tsinghua University, Beijing, China, in 2017.
He is currently pursuing the Ph.D. degree in the
Department of Automation, Tsinghua University, Beijing, China.
\end{IEEEbiography}

\begin{IEEEbiography}[{\includegraphics[width=1in,height=1.25in,clip,keepaspectratio]{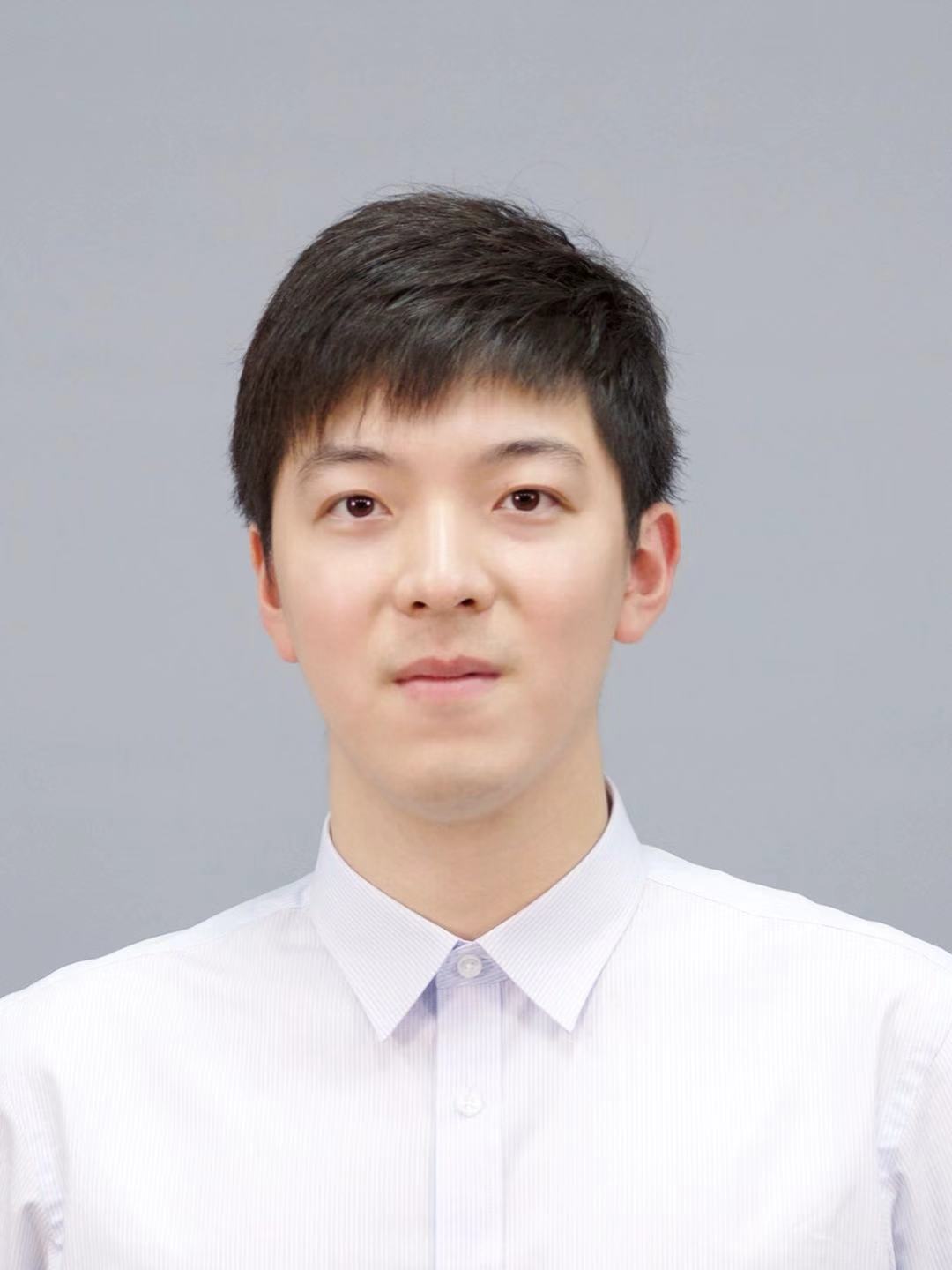}}]{Weilin Wan}
received the B.S. degree in computer science from the University of Washington, Seattle, USA, in 2018. He is working toward his PhD degree in the Department of Computer Science, the University of Hong Kong. His research interests include computer graphics, machine learning, and robotics.
\end{IEEEbiography}

\begin{IEEEbiography}[{\includegraphics[width=1in,height=1.25in,clip,keepaspectratio]{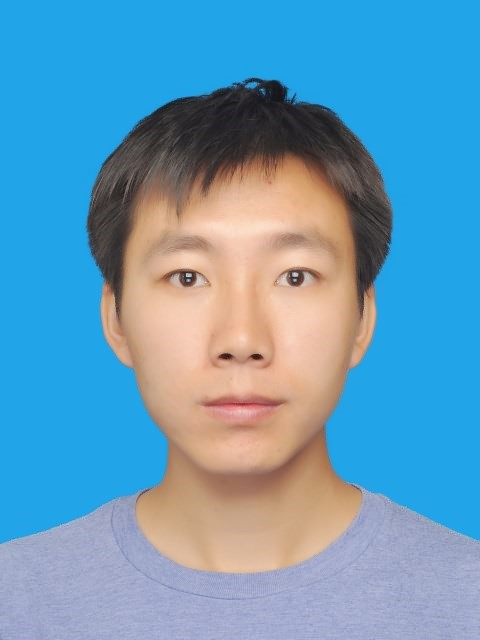}}]{Tao Yu}
is a post-doctoral researcher at Tsinghua University. He received the B.S. degree in Measurement
and Control from Hefei University of Technology,
China, in 2012, and the Ph.D. degree
in instrumental science from
Beihang University, China.
His current research
interests include computer vision and computer
graphics.
\end{IEEEbiography}

\begin{IEEEbiography}[{\includegraphics[width=1in,height=1.25in,clip,keepaspectratio]{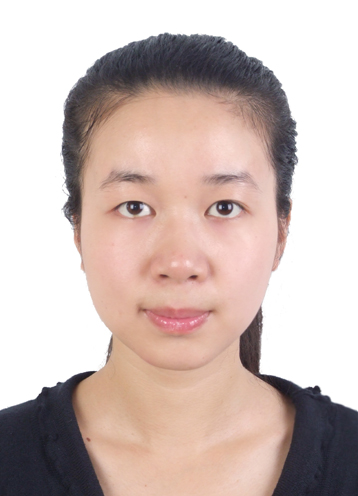}}]{Lingjie Liu}
is a post-doctoral researcher at the Graphic, Vision \& Video group of Max Planck Institute for Informatics in Saarbrücken, Germany. She received the B.E. degree from the Huazhong University of Science and Technology in 2014 and the PhD degree from the University of Hong Kong in 2019. Her research interests include 3D reconstruction, human performance capture and video synthesis. She has received Hong Kong PhD Fellowship Award (2014) and Lise Meitner Fellowship Award (2019).
\end{IEEEbiography}

\begin{IEEEbiography}[{\includegraphics[width=1in,height=1.25in,clip,keepaspectratio]{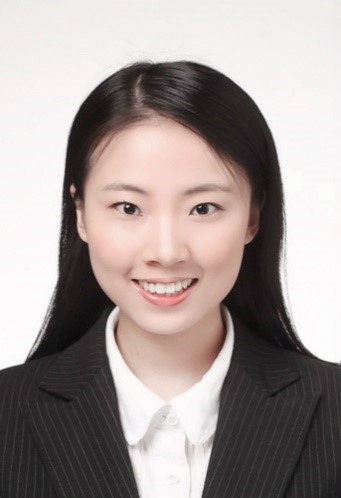}}]{Lu Fang}
is currently an Associate Professor
at Tsinghua University. She received her Ph.D
in Electronic and Computer Engineering from
HKUST in 2011, and B.E. from USTC in 2007,
respectively. Dr. Fang’s research interests include
image/video processing, vision for intelligent
robot, and computational photography. Dr.
Fang serves as TC member in Multimedia Signal
Processing Technical Committee (MMSP-TC) in
IEEE Signal Processing Society.
\end{IEEEbiography}

\begin{IEEEbiography}[{\includegraphics[width=1in,height=1.25in,clip,keepaspectratio]{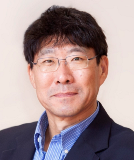}}]{Wenping Wang}
 is currently a Chair Professor at the Department of Computer Science, the University of Hong Kong. His research interests include computer graphics, visualization, and geometric computing. He has made fundamental research contributions in collision detection, shape modeling and analysis, mesh generation, and architectural geometry. He is journal associate editor of Computer Aided Geometric Design (CAGD), Computers and Graphics (CAG), IEEE Transactions on Visualization and Computer Graphics (TVCG, 2008-2012), Computer Graphics Forum (CGF), IEEE Computer Graphics and Applications, and IEEE Transactions on Computers. He received the Outstanding Researcher Award of the University of Hong Kong in 2013. He received John Gregory Award in 2017 for contributions in geometric modeling and computing. He is an IEEE Fellow.
\end{IEEEbiography}

\begin{IEEEbiography}[{\includegraphics[width=1in,height=1.25in,clip,keepaspectratio]{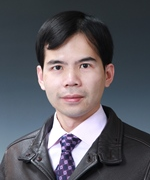}}]{Yebin Liu}
is currently an associate professor at
Tsinghua University. He received the B.E. degree
from the Beijing University of Posts and
Telecommunications, China, in 2002, and the
PhD degree from the Automation Department,
Tsinghua University, Beijing, China, in 2009. He
was a research fellow in the Computer Graphics
Group of the Max Planck Institute for Informatik,
Germany, in 2010. His research areas include
computer vision, computer graphics and computational
photography.
\end{IEEEbiography}

\vfill






\end{document}